\documentclass[11pt,a4paper]{article}
\usepackage[hyperref]{acl2019}
\usepackage{lscape} 
\usepackage{times}
\usepackage{latexsym}
\usepackage{amsmath}
\usepackage{bbm}
\usepackage{dsfont}
\usepackage{amssymb}
\usepackage{url}
\usepackage{tabularx}
\usepackage{multirow}
\usepackage{cleveref}
\crefformat{section}{\S#2#1#3} 
\crefformat{subsection}{\S#2#1#3}
\crefformat{subsubsection}{\S#2#1#3}
\usepackage{graphicx}
\usepackage{tikz}
\usetikzlibrary{fit,positioning}
\usetikzlibrary{bayesnet}
\usetikzlibrary{arrows}
\usetikzlibrary{patterns}
\usepackage{comment}
\usepackage{subcaption}
\tikzstyle{det} = [latent, diamond]
\usepackage{tabularx}

\usepackage[noend]{algpseudocode}
\usepackage{algorithm}
\algrenewcommand\algorithmicindent{1em}
\usepackage[font=footnotesize,labelfont=bf]{caption}
\makeatletter
\algrenewcommand\ALG@beginalgorithmic{\small}
\makeatother
\usepackage{booktabs}
\usepackage{url}
\newcommand\blfootnote[1]{%
  \begingroup
  \renewcommand\thefootnote{}\footnote{#1}%
  \addtocounter{footnote}{-1}%
  \endgroup
}
\aclfinalcopy 

\usepackage{tikz-qtree}
\newcommand{\sentence}{\boldsymbol{x}}
\newcommand{\boldx}{\mathbf{x}}
\newcommand{\tree}{\boldsymbol{t}}

\newcommand{\boldy}{\mathbf{y}}
\newcommand{\boldw}{\mathbf{w}}

\newcommand{\boldz}{\mathbf{z}}

\newcommand{\boldE}{\mathbf{E}}
\newcommand{\boldb}{\mathbf{b}}

\newcommand{\boldq}{\mathbf{q}}
\newcommand{\boldp}{\mathbf{p}}

\newcommand{\boldu}{\mathbf{u}}

\newcommand{\boldh}{\mathbf{h}}

\newcommand{\mcN}{\mathcal{N}}
\newcommand{\mcG}{\mathcal{G}}
\newcommand{\mcR}{\mathcal{R}}
\newcommand{\mcM}{\mathcal{M}}
\newcommand{\bpi}{\boldsymbol{\pi}}

\newcommand{\mcT}{\mathcal{T}}
\newcommand{\mcP}{\mathcal{P}}

\newcommand{\E}{\mathbb{E}}
\newcommand{\reals}{\mathbb{R}}

\DeclareMathOperator*{\KL}{KL}
\DeclareMathOperator*{\ELBO}{ELBO}

\DeclareMathOperator*{\argmax}{argmax}

\DeclareMathOperator*{\relu}{ReLU}

\DeclareMathOperator*{\MLP}{MLP}

\newcommand{\given}{\,|\,}
\newcommand{\param}{;}

\title{Compound Probabilistic Context-Free Grammars \\ for Grammar Induction}

\author{Yoon Kim  \\
  Harvard University \\
  Cambridge, MA, USA \\
  \texttt{\fontsize{11pt}{12pt}\selectfont yoonkim@seas.harvard.edu} \\ \And
  Chris Dyer \\
  DeepMind \\
  London, UK \\
  \texttt{\fontsize{11pt}{12pt}\selectfont cdyer@google.com} \\
  \And
  Alexander M. Rush \\
  Harvard University \\
  Cambridge, MA, USA \\
  \texttt{\fontsize{11pt}{12pt}\selectfont srush@seas.harvard.edu} \\}

\date{}

\begin{document}
\maketitle
\begin{abstract}
We study a formalization of the grammar induction problem that models sentences as being generated by a compound probabilistic context-free grammar. In contrast to traditional formulations which learn a single stochastic grammar, our grammar's rule probabilities are modulated by a per-sentence continuous latent variable, which induces marginal dependencies beyond the traditional context-free assumptions. 
Inference in this grammar is performed by collapsed variational inference, in which an amortized variational posterior is placed on the continuous variable, and the latent trees are marginalized out with dynamic programming. Experiments on English and Chinese show the effectiveness of our approach compared to recent state-of-the-art methods when evaluated on unsupervised parsing.
\end{abstract}
\section{Introduction}

 \blfootnote{\noindent \hspace{-6mm} Code: \url{https://github.com/harvardnlp/compound-pcfg}}
Grammar induction is the task of inducing hierarchical syntactic structure from data. Statistical approaches to grammar induction require specifying a probabilistic grammar (e.g. formalism, number and shape of rules), and fitting its parameters through optimization.  Early work found that it was difficult to induce probabilistic context-free grammars (PCFG) from natural language data through direct methods, such as optimizing the log likelihood with the EM algorithm \cite{lari1990scfg,carroll1992two}. While the reasons for the failure are manifold and not completely understood, two major potential causes are the ill-behaved optimization landscape and the overly strict independence assumptions of PCFGs.
More successful approaches to grammar induction have thus resorted to carefully-crafted auxiliary objectives \cite{klein2002ccm}, priors or non-parametric models \cite{kurihara2006var,johnson2007pcfg,liang2007pcfg,wang2013collapsed}, and manually-engineered features \cite{huang2012,golland2012} to encourage the desired structures to emerge.

We revisit these aforementioned issues in light of advances in model parameterization and inference. First, contrary to common wisdom, we find that parameterizing a PCFG's rule probabilities with neural networks over distributed representations makes it possible to induce linguistically meaningful grammars  by simply optimizing log likelihood. While the optimization problem remains non-convex, recent work suggests that there are optimization benefits afforded by over-parameterized models \cite{arora2018opt,xu2018over,du2019opt}, and we indeed find that this neural PCFG is significantly easier to optimize than the traditional PCFG. Second, this factored parameterization makes it straightforward to incorporate side information into 
rule probabilities through a sentence-level continuous latent vector, which effectively allows different contexts in a derivation to coordinate. In this compound PCFG---continuous mixture of PCFGs---the context-free assumptions hold conditioned on the latent vector but not unconditionally, thereby obtaining longer-range dependencies within a tree-based generative process. 

To utilize this approach, we need to efficiently optimize the log marginal likelihood of observed sentences. While compound PCFGs break efficient inference, if the latent vector is known the distribution over trees reduces to a standard PCFG. This property allows us to perform grammar induction using a collapsed approach where the latent trees are marginalized out exactly with dynamic programming. To handle the latent vector, we employ standard amortized inference using reparameterized samples from a variational posterior approximated from an inference network \cite{kingma2014vae,rezende2014vae}. 

On standard benchmarks for English and Chinese, the proposed approach is found to perform favorably against recent neural approaches to unsupervised parsing \cite{shen2018nlm,shen2019ordered,drozdov2018latent,kim2019urnng}.

\section{Probabilistic Context-Free Grammars}

We consider context-free grammars (CFG)
consisting of a  5-tuple $\mcG = (S, \mcN, \mcP, \Sigma, \mcR)$ where $S$ is the distinguished start symbol, $\mcN$ is a finite set of nonterminals, $\mcP$ is a finite set of preterminals,\footnote{Since we will be inducing a grammar directly from words,  $\mcP$ is roughly the set of part-of-speech tags and $\mcN$ is the set of constituent labels. However, to avoid issues of label alignment, evaluation is only on the tree topology.} $\Sigma$ is a finite set of terminal symbols, and $\mcR$ is a finite set of rules of the form, 
\begin{align*}
    S \to A, & & A \in \mcN \\
    A \to B\ C, & &A \in \mcN, \, \,\, B,C \in \mcN \cup \mcP \\
    T \to w, & &T \in \mcP, \,\, \, w \in \Sigma. 
\end{align*}
A probabilistic context-free grammar (PCFG) consists of a grammar $\mcG$ and rule probabilities $\boldsymbol{\pi} = \{\pi_r\}_{r \in \mcR}$ such that $\pi_r$ is the probability of the rule $r$. Letting $\mcT_\mcG$ be the set of all parse trees of $\mcG$, a PCFG
defines a probability distribution over $\tree \in \mcT_\mcG$ via $p_{\bpi}(\tree) = \prod_{r \in \tree_\mcR} \pi_r$ where $\tree_\mcR$ is the set of rules used in the derivation of $\tree$. It also defines a distribution over string of terminals 
$ \sentence \in \Sigma^\ast$ via 
\begin{align*}p_{\bpi}(\sentence) = \sum_{\tree \in \mcT_\mcG(\sentence)} p_{\bpi}(\tree), 
\end{align*}
where $\mcT_\mcG(\sentence) = \{ \tree \, \vert \, \textsf{yield}(\tree) = \sentence\}$, i.e. the set of trees $\tree$ such that $\tree$'s leaves are $\sentence$. We will slightly abuse notation and use  
\begin{align*}
    p_{\bpi}(\tree_{} \given \sentence)  \triangleq  \frac{\mathds{1}[\textsf{yield}(\tree) = \sentence]p_{\bpi}(\tree)}{p_{\bpi}(\sentence)} 
\end{align*} to denote the posterior distribution over the unobserved latent trees  given the observed sentence $\sentence$, where $\mathds{1}[\cdot]$ is the indicator function.\footnote{Therefore  when used in the context of a posterior distribution conditioned on a sentence $\sentence$, the variable $\tree$ does not include the leaves $\sentence$ and only refers to the unobserved nonterminal/preterminal symbols.}

    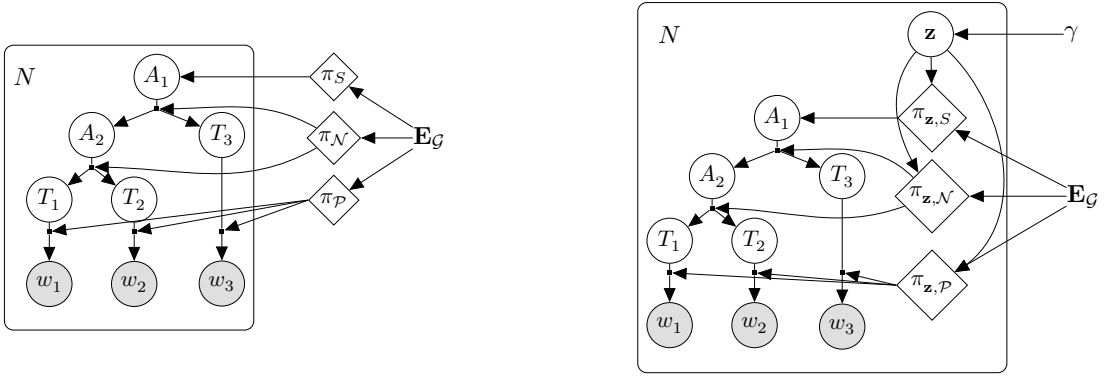
\begin{figure*}
    \hspace{10mm}
\begin{subfigure}{0.4\textwidth}
\centering
\scalebox{0.85}{

\begin{tikzpicture}

\tikzstyle{factor} = [rectangle, fill=black,minimum size=2pt, inner
sep=0pt, node distance=0.4]

\node[latent] (z2) {$A_1^{}$};%
\node[latent, below left=0.4cm and 0.5cm of z2] (z3) {$A_2^{}$};
\node[latent, below right=0.4cm and 0.5cm of z2] (z4) {$T_3^{}$};
\node[latent, below left=0.5cm and 0.15cm of z3] (z5) {$T_1^{}$};
 \node[latent, below right=0.5cm and 0.15cm of z3] (z6) {$T_2^{}$};
     \node[obs, below = 0.6cm of z5 ](w1) {$w_1^{}$};
    \node[obs, below = 0.6cm of z6 ](w2) {$w_2^{}$};
      \node[obs, below = 1.60cm of z4 ](w3) {$w_3^{}$};

\node[det, right= 2.0cm of z2] (theta1) {$\pi_S$};
\node[det, below =0.15cm of theta1] (theta2) {$\pi_\mcN$};
\node[det, below =0.15cm of theta2] (theta3) {$\pi_\mcP$};
\node[const, right= 0.8cm of theta2] (phi) {$\boldE_\mcG$};
   \node[const, left =1.5cm of z2 ] (n) {$N$};

\plate {plate1} {(n)(z2)(z3)(z4)(z5)(z6)(w1)(w2)(w3)}{};

  \factor[below=(0.1cm) of z2]     {mid}     {} {} {} ; %
 \factor[below=(0.1cm) of z3]     {mid2}     {} {} {} ; %

  \factoredge {z2} {mid} {z3,z4};
    \factoredge {z3} {mid2} {z5,z6};

\factor[below=(0.1cm) of z5]     {mid3}     {} {} {} ; %
 \factor[below=(0.1cm) of z6]     {mid4}     {} {} {} ; %
 \factor[below=(1.1cm) of z4]     {mid5}     {} {} {} ; %

\factoredge {z5} {mid3} {w1};
\factoredge {z6} {mid4} {w2};
\factoredge {z4} {mid5} {w3};

\edge {theta1} {z2};

\draw [->] (theta2) to [out=150,in=0] (mid);
\draw [->] (theta2) to [out=210,in=0] (mid2);

\edge {phi} {theta1,theta2,theta3};
\edge {theta3.west} {mid5};
\draw [->] (theta3.west) to  (mid3);
\draw [->] (theta3.west) to  (mid4);
 \end{tikzpicture} }
 \end{subfigure}
\hspace{20mm}
 \begin{subfigure}{0.5\textwidth}
 \scalebox{0.85}{
 \begin{tikzpicture}
\tikzstyle{factor} = [rectangle, fill=black,minimum size=2pt, inner
sep=0pt, node distance=0.4]
\node[latent] (z2) {$A_{1}^{}$};%
\node[latent, below left=0.4cm and 0.5cm of z2] (z3) {$A_{2}^{}$};%
\node[latent, below right=0.4cm and 0.5cm of z2] (z4) {$T_{3}^{}$};%
\node[latent, below left=0.5cm and 0.15cm of z3] (z5) {$T_{1}^{}$};%
 \node[latent, below right=0.5cm and 0.15cm of z3] (z6) {$T_{2}^{}$};%
     \node[obs, below = 0.6cm of z5 ](w1) {$w_1^{}$};%
    \node[obs, below = 0.6cm of z6 ](w2) {$w_2^{}$};%
      \node[obs, below = 1.63cm of z4 ](w3) {$w_3^{}$};%
      \node[latent, above right = 0.8cm and 1.87cm of z2] (prior) {$\boldz^{}$};
       \node[const, right= 1.7cm of prior] (gamma) {$\gamma$};
        \node[const, right= 0.5cm of prior] (c) {\color{white} c};
\node[det, right= 1.5cm of z2] (theta1) {$\pi_{\boldz,S}^{}$};
\node[det, below =0.1cm of theta1] (theta2) {$\pi_{\boldz, \mcN}^{}$};
\node[det, below =0.25cm of theta2] (theta3) {$\pi_{\boldz,\mcP}^{}$};
\node[const, right= 1.5cm of theta2] (phi) {$\boldE_\mcG$};
\node[const, right= 0.5cm of theta2] (dummy) {$$};
   \node[const, left =3.5cm of prior ] (n) {$N$};
\plate {plate1} {(n)(z2)(z3)(z4)(z5)(z6)(w1)(w2)(w3)(prior)(theta1)(theta2)(theta3)(c)}{};
  \factor[below=(0.1cm) of z2]     {mid}     {} {} {} ; %
 \factor[below=(0.1cm) of z3]     {mid2}     {} {} {} ; %

  \factoredge {z2} {mid} {z3,z4};
    \factoredge {z3} {mid2} {z5,z6};

\factor[below=(0.1cm) of z5]     {mid3}     {} {} {} ; %
 \factor[below=(0.1cm) of z6]     {mid4}     {} {} {} ; %
 \factor[below=(1.1cm) of z4]     {mid5}     {} {} {} ; %

\factoredge {z5} {mid3} {w1};
\factoredge {z6} {mid4} {w2};
\factoredge {z4} {mid5} {w3};

\edge {prior} {theta1};

\edge {theta1} {z2};

\draw [->] (theta2) to [out=135,in=0] (mid);
\draw [->] (theta2) to [out=200,in=0] (mid2);
\draw [->] (prior) to [out=-130,in=120] (theta2);
\draw [->] (prior) to [out=-44,in=30] (theta3);
\edge {phi} {theta1,theta2,theta3};
\edge {theta3.west} {mid5};
\draw [->] (theta3.west) to  (mid3);
\draw [->] (theta3.west) to  (mid4);

\edge {gamma} {prior};

 \end{tikzpicture}
 }
 \end{subfigure}

 \caption{A graphical model-like diagram for the neural PCFG (left) and the compound PCFG (right) for an example tree structure. 
 In the above, $A_1^{},A_2^{} \in \mcN$ are nonterminals, $T_1^{},T_2^{}, T_3^{}\in \mcP$ are preterminals, $ w_1^{}, w_2^{}, w_3^{} \in \Sigma$ are terminals. In the neural PCFG, the global rule probabilities $\bpi = \pi_S \cup \pi_\mcN \cup \pi_\mcP$ are the output from a neural net run over the symbol embeddings $\boldE_\mcG$, where $\pi_\mcN$ are the set of rules with a nonterminal on the left hand side ($\pi_S$ and $\pi_\mcP$ are similarly defined).
 In the compound PCFG, we have per-sentence rule probabilities $\bpi_\boldz = \pi_{\boldz,S} \cup \pi_{\boldz, \mcN} \cup \pi_{\boldz, \mcP}$ obtained from running a neural net over a random vector $\boldz^{}$ (which varies across sentences) and global symbol embeddings $\boldE_\mcG$. In this case, the context-free assumptions hold conditioned on $\boldz^{}$, but they do not hold unconditionally: e.g. when conditioned on $\boldz^{}$ and $A_{2}$, the variables $A_{1}$ and $T_{1}$ are independent; however when conditioned on just $A_{2}$, they are not independent due to the dependence path through $\boldz^{}$. Note that the rule probabilities  are random variables in the compound PCFG but deterministic variables in the neural PCFG.}
  \label{fig:gm}

\end{figure*}

\subsection{Parameterization} The standard way to parameterize a PCFG is to simply associate a scalar to each rule $\pi_r$ with the constraint that they form valid probability distributions, i.e. each nonterminal is associated 
with a fully-parameterized categorical distribution over its rules. 
This \textit{direct parameterization} is algorithmically convenient since the M-step in the EM algorithm \cite{dempster77em} has a closed form. However, there is a 
long history of work showing that it is difficult to learn meaningful grammars from natural language data with this parameterization \cite{carroll1992two}.\footnote{In preliminary experiments we were indeed unable to learn linguistically meaningful grammars with this PCFG.} Successful approaches to unsupervised parsing have therefore modified the model/learning objective by guiding potentially unrelated rules to behave similarly.

Recognizing that sharing among rule types is beneficial, we propose a \textit{neural parameterization} where rule probabilities are based on distributed representations. We associate embeddings with each symbol, introducing input embeddings $\boldw_N$ for each symbol $N$ on the left side of a rule (i.e. $N \in \{S\} \cup  \mcN \cup \mcP$).
For each rule type $r$, $\pi_r$ is parameterized as follows,    
\begin{align*}
   \pi_{S \to A} &= \frac{\exp(\boldu^\top_A \,  f_1(\boldw_S)  + b_A)}{\sum_{A' \in \mcN}\exp(\boldu^\top_{A'} \,  f_1(\boldw_S) + b_{A'} ) },  \\
    \pi_{A \to BC} &= \frac{\exp(\boldu^\top_{BC} \, \boldw_{A} + b_{BC})}{\sum_{B'C' \in \mcM} \exp(\boldu^\top_{B'C'} \, \boldw_{A} + b_{B'C'})}, \\ 
    \pi_{T \to w} &= \frac{\exp(\boldu^\top_w \,  f_2(\boldw_T) + b_w)}{\sum_{w' \in \Sigma}\exp(\boldu^\top_{w'} \,  f_2(\boldw_T)  + b_{w'}) } ,
\end{align*}
where $\mcM$ is the product space $(\mcN \cup \mcP) \times (\mcN \cup \mcP)$, and $f_1, f_2$ are MLPs with two residual layers. Note that we do not use an MLP for rules of the type $\pi_{A \to BC}$, as it did not empirically improve results. See \cref{lab:model} for the full parameterization. Going forward, we will use $\boldE_\mcG = \{\boldw_U \, \vert \, U \in \{S\} \cup \mcN \cup \mcP \} \cup \{\boldu_V \, \vert \, V \in \mcN \cup \mathcal{M} \cup \Sigma \}$ to denote the set of input/output symbol embeddings for grammar $\mcG$, and $\lambda$ to refer to the parameters of the neural network $f_1, f_2$ used to obtain the rule probabilities. A graphical model-like illustration of the neural PCFG is shown in Figure~\ref{fig:gm} (left).

It is clear that the neural parameterization does not change the underlying probabilistic assumptions. The difference between the two is analogous to the difference between count-based vs. feed-forward neural language models, where feed-forward neural language models make the same Markov assumptions as the count-based models but are able to take advantage of shared, distributed representations.  

\section{Compound PCFGs}

A compound probability distribution \cite{robbins1951compound} is a distribution whose parameters are themselves random variables. These distributions generalize mixture models to the continuous case, for example in factor analysis which assumes the following generative process,
\begin{align*}
    \boldz^{} \sim \text{N}(\mathbf{0}, \mathbf{I}), &&\boldx \sim \text{N}(\mathbf{W}\boldz, \boldsymbol{\Sigma}). 
\end{align*}
Compound distributions provide the ability to model rich generative processes, but marginalizing over the latent parameter can be computationally intractable unless conjugacy can be exploited.

In this work, we study compound probabilistic context-free grammars whose distribution over trees arises from the following generative process: we first obtain rule probabilities via
\begin{align*}
   \boldz^{} \sim p_\gamma(\boldz), & & \bpi_\boldz = f_\lambda(\boldz^{}, \boldE_\mcG), 
\end{align*}
where $p_\gamma(\boldz)$ is a prior with parameters $\gamma$ (spherical Gaussian in this paper), and $f_\lambda$ is a neural network that concatenates the input symbol embeddings with $\boldz^{}$ and outputs the sentence-level rule probabilities $\bpi_\boldz$,
\begin{align*}
   \pi_{\boldz, S \to A}^{} & \propto \exp(\boldu^\top_A \,  f_1([\boldw_S \param \boldz^{}]) + b_A), \\
    \pi_{\boldz, A \to B C }^{}  &\propto \exp(\boldu^\top_{BC} \, [\boldw_{A} \param \boldz^{}] + b_{BC}), \\ 
    \pi_{\boldz, T \to w}^{} &\propto \exp(\boldu^\top_w \,  f_2([\boldw_T \param \boldz^{}]) + b_w), 
\end{align*}
where $[\boldw \param \boldz]$ denotes vector concatenation.
Then a tree/sentence is sampled from a PCFG with rule probabilities given by  $\bpi_\boldz$,
\begin{align*}
         \tree^{} \sim \text{PCFG}(\bpi_\boldz), && \sentence^{} = \textsf{yield}(\tree^{}). 
\end{align*}
This can be viewed as a continuous mixture of PCFGs, or alternatively, a Bayesian PCFG with a prior on sentence-level rule probabilities parameterized by $\boldz, \lambda,\boldE_\mcG$.\footnote{Under the Bayesian PCFG view, $p_\gamma(\boldz)$ is a distribution over $\boldz$ (a subset of the prior), and is thus a hyperprior.} Importantly, under this generative model the context-free assumptions hold \emph{conditioned on $\boldz^{}$}, but they do not hold unconditionally. This is shown  in Figure~\ref{fig:gm} (right) where there is a dependence path through $\boldz^{}$ if it is not conditioned upon. Compound PCFGs give rise to a marginal distribution over parse trees $\tree$ via
\begin{align*}
p_\theta(\tree) =  \int p(\tree \given \boldz) p_\gamma(\boldz) \, \mathrm{d}\boldz, 
\end{align*} where $p_\theta(\tree \given \boldz) = \prod_{r \in \tree_{\mcR }} \pi_{ \boldz, r}$. The subscript in  $\pi_{\boldz, r}$ denotes the fact that the rule probabilities depend on $\boldz$. Compound PCFGs are clearly more expressive than PCFGs as each sentence has its own set of rule probabilities. However, it still assumes a tree-based generative process, making it possible to learn latent tree structures. 

One motivation for the compound PCFG is that simple, unlexicalized grammars (such as the PCFG we have been working with) are unlikely to represent an adequate model of natural language, although they do facilitate efficient learning and inference.\footnote{A piece of evidence for the misspecification of unlexicalized first-order PCFGs as a statistical model of natural language is that if one pretrains such a PCFG on supervised data and continues training with the unsupervised objective (i.e. log marginal likelihood), the resulting grammar deviates significantly from the supervised initial grammar while the log marginal likelihood improves \cite{johnson2007pcfg}. Similar observations have been made for part-of-speech induction with Hidden Markov Models \cite{merialdo:1994}.} We can in principle model richer dependencies through vertical/horizontal Markovization \cite{johnson1998,klein2003} and lexicalization \cite{collins1997three}. However such dependencies complicate training due to the rapid increase in the number of rules. Under this view, we can interpret the compound PCFG as a restricted version of some lexicalized, higher-order PCFG where a child can depend on structural and lexical context through a shared latent vector.\footnote{Note that the compound ``PCFG" is a slight misnomer because the model is no longer context-free in the usual sense. Another interpretation of the model is to view it as a vectorized version of \emph{indexed} grammars \cite{aho1968}, which extend CFGs by augmenting nonterminals with additional index strings that may be inherited or modified during derivation. Compound PCFGs instead equip nonterminals with a continuous vector that is always inherited.}  We hypothesize that this dependence among siblings is especially useful in grammar induction from words, where (for example) if we know that  \textsf{watched} is used as a verb then the noun phrase is likely to be \textsf{a movie}.

In contrast to the usual Bayesian treatment of PCFGs which places priors on global rule probabilities \cite{kurihara2006var,johnson2007pcfg,wang2013collapsed}, the compound PCFG assumes a prior on local, sentence-level rule probabilities. It is therefore closely related to the Bayesian grammars studied by \citet{cohen2009logistic} and \citet{cohen2009shared}, who also sample local rule probabilities from a logistic normal prior for training dependency models with valence (DMV) \cite{klein2004syntax}. 

\subsection{Inference in Compound PCFGs} The expressivity of compound PCFGs comes at a significant challenge in learning and inference. 
Letting $\theta = \{\boldE_{\mcG}, \lambda \}$ be the parameters of the generative model, we would like to maximize the log marginal likelihood of the observed sentence $\log p_\theta(\sentence)$. In the neural PCFG the log marginal likelihood 
\[\log p_\theta(\sentence) = \log \sum_{\tree \in \mcT_\mcG(\sentence)} p_\theta(\tree),\]
can be obtained by summing out the latent tree structure using the inside algorithm \cite{baker1979io}, which is differentiable and thus amenable to gradient-based optimization.\footnote{In the context of the EM algorithm, directly performing gradient ascent on the log marginal likelihood is equivalent to performing an exact E-step (with the inside-outside algorithm) followed by a gradient-based M-step, i.e. $\nabla_\theta \log p_\theta(\sentence) = \E_{p_\theta(\tree \given \sentence)}[\nabla_\theta \log p_\theta(\tree)]$ \cite{salak2003,kirk2010,eisner2016}.}
In the compound PCFG, the log marginal likelihood is given by, 
\begin{align*}
    \log p_\theta(\sentence) &= \log \Big( \int  p_\theta(\sentence \given \boldz) p_\gamma(\boldz)\, \mathrm{d}\boldz \Big) \\
    &=\log \Big( \int \sum_{\tree \in \mcT_\mcG(\sentence)}  p_\theta(\tree \given \boldz) p_\gamma(\boldz)\, \mathrm{d}\boldz \Big).  
\end{align*} 
Notice that while the integral over $\boldz$ makes this quantity intractable, when we condition on $\boldz$, we can tractably perform the inner summation to obtain $p_\theta(\sentence \given \boldz)$ using the inside algorithm. We therefore resort to collapsed amortized variational inference. We first obtain a sample $\boldz$ from a variational posterior distribution (given by an amortized inference network), then perform the inner marginalization conditioned on this sample. The evidence lower bound $ \ELBO(\theta, \phi \param \sentence)$ is then, 
\begin{align*}
\E_{q_\phi(\boldz \given \sentence)}[\log  p_\theta(\sentence \given \boldz)]  
- \KL[q_\phi(\boldz \given \sentence)\, \Vert \, p_\gamma(\boldz)],  
\end{align*}
and  we can calculate $p_\theta(\sentence \given \boldz)$ given a sample $\boldz$ from a variational posterior $q_\phi(\boldz \given \sentence)$.
For the variational family we use a diagonal Gaussian where the mean/log-variance vectors are given by an affine layer over max-pooled hidden states from an LSTM over $\sentence$. We can obtain low-variance estimators for $\nabla_{\theta,\phi}\ELBO(\theta, \phi \param \sentence)$ by using the reparameterization trick for the expected reconstruction likelihood and the analytical expression for the KL term \cite{kingma2014vae}.  

We remark that under the Bayesian PCFG view, since the parameters of the prior (i.e. $\theta$) are estimated from the data, our approach can be seen as an instance of empirical Bayes \cite{robbins1956empirical}.\footnote{See \citet{berger1985bayes} (chapter 4), \citet{zhang2003compound}, and \citet{cohen2016bayesian} (chapter 3)  for further discussion on compound models and empirical Bayes.} 

\subsection{MAP Inference}
After training, we are interested in comparing the learned trees against an annotated treebank. This requires inferring the most likely tree given a sentence, i.e. $\argmax_{\tree}\,\, p_\theta(\tree \given \sentence)$.
For the neural PCFG we can obtain the most likely tree by using the Viterbi version of the inside algorithm (CKY algorithm). For the compound PCFG, the $\argmax$ is intractable to obtain exactly, and hence we estimate it with the following approximation, 
\begin{align*}
  &  \argmax_{\tree} \,\, \int p_\theta(\tree \given \sentence, \boldz) p_\theta(\boldz \given \sentence) \, \mathrm{d}\boldz \\
    & =  \argmax_{\tree} \,\,  p_\theta\big(\tree \given \sentence,\boldsymbol{\mu}_\phi(\sentence) \big),
\end{align*}
where $\boldsymbol{\mu}_\phi(\sentence)$ is the mean vector from the inference network.
The above approximates the true  posterior $p_\theta(\boldz \given \sentence)$ with $\delta(\boldz -  \boldsymbol{\mu}_\phi(\sentence))$, the Dirac delta function at the mode of the variational posterior.\footnote{Since $p_\theta(\tree \given \sentence, \boldz)$ is continuous with respect to $\boldz$, we have $
\int p_\theta(\tree \given \sentence, \boldz)\delta(\boldz -  \boldsymbol{\mu}_\phi(\sentence)) \, \mathrm{d}\boldz = p_\theta\big(\tree \given \sentence,\boldsymbol{\mu}_\phi(\sentence) \big).$}
This quantity is tractable as in the PCFG case. Other approximations are possible: for example we could use $q_\phi(\boldz \given \sentence)$ as an importance sampling distribution to estimate the first integral. However we found the above approximation to be efficient and effective in practice.

\section{Experimental Setup}

\subsection{Data}
We test our approach on the Penn Treebank (PTB) \cite{marcus1993ptb} with the standard splits (2-21 for training, 22 for validation, 23 for test) and the same preprocessing as in recent works \cite{shen2018nlm,shen2019ordered}, where we discard punctuation, lowercase all tokens, and take the top 10K most frequent words as the vocabulary. This setup is more challenging than traditional setups, which usually experiment on shorter sentences and use gold part-of-speech tags. 

We further experiment on Chinese with version 5.1 of the Chinese Penn Treebank (CTB) \cite{xue2005ctb}, with the same splits as in \citet{chen2014fast}. On CTB we also remove punctuation and keep the top 10K word types.

\subsection{Hyperparameters} Our PCFG uses 30 nonterminals and 60 preterminals, with 256-dimensional symbol embeddings. The compound PCFG uses 64-dimensional latent vectors. The bidirectional LSTM inference network has a single layer with 512 dimensions, and the mean and the log variance vector for $q_\phi(\boldz \given \sentence)$ are given by max-pooling the hidden states of the LSTM and passing it through an affine layer. Model parameters are initialized with Xavier uniform initialization. 
For training we use Adam \cite{kingma2015adam} with $\beta_1$ = 0.75, $\beta_2 = 0.999$ and learning rate of 0.001, with a maximum gradient norm limit of 3. We train for 10 epochs with batch size equal to 4. We employ a curriculum learning strategy \cite{bengio2009curr} where we train only on sentences of length up to 30 in the first epoch, and increase this length limit by 1 each epoch. Similar curriculum-based strategies have used in the past for grammar induction \cite{spitkovsky2012three}.
 During training we perform early stopping based on validation perplexity.\footnote{However, we used $F_1$ against validation trees on PTB to select some hyperparameters (e.g. grammar size), as is sometimes done in grammar induction. Hence our PTB results are arguably not fully unsupervised in the strictest sense of the term.  The hyperparameters of the PRPN/ON baselines are also tuned using validation $F_1$ for fair comparison.} Finally, to mitigate against overfitting to PTB, experiments on CTB utilize the same hyperparameters from PTB.

\subsection{Baselines and Evaluation}
While we induce a full stochastic grammar (i.e. a distribution over symbolic rewrite rules) in this work, directly assessing the learned grammar is itself nontrivial. As a proxy, we adopt the usual approach and instead evaluate the induced grammar as an unsupervised parsing system. 
However, even in this setting we observe that there is enough variation  across prior work on  to render a meaningful comparison difficult. 

In particular, some important dimensions along which prior works vary include, (1) input data: earlier work on  generally assumed gold (or induced) part-of-speech tags \cite{klein2004syntax,smith2004,bod2006subtrees,snyder2009unsup}, while more recent works induce grammar directly from words \cite{spitkovsky2013breaking,shen2018nlm}; (2) use of punctuation: even within papers that induce parse trees directly from words, some papers employ heuristics based on punctuation as punctuation is usually a strong signal for start/end of constituents  \cite{seginer2007unsup,ponvert2011simple,spitkovsky2013breaking}, some train  with punctuation \cite{jin2018depth,drozdov2018latent,kim2019urnng}, while others discard punctuation altogether for training \cite{shen2018nlm,shen2019ordered}; (3) train/test data: some works do not explicitly separate out train/test sets  \cite{reichart2010zoom,golland2012} while some do  \cite{huang2012,parikh2014spectral,htut2018grammar}. Maintaining train/test splits is less of an issue for unsupervised structure learning, however in this work we follow the latter and separate train/test data. (4) evaluation: for unlabeled $F_1$, almost all works ignore punctuation (even approaches that use punctuation during training typically ignore them during evaluation), but there is some variance in discarding trivial spans (width-one and sentence-level spans) and using corpus-level versus sentence-level $F_1$.\footnote{Corpus-level $F_1$ calculates precision/recall at the corpus level to obtain $F_1$, while sentence-level $F_1$ calculates $F_1$ for each sentence and averages across the corpus.} In this paper we discard trivial spans and evaluate on sentence-level $F_1$ per recent work \cite{shen2018nlm,shen2019ordered}.
\begin{table}[]
\small
    \centering
    \begin{tabular}{l@{\hskip 1mm}  r@{\hskip 3mm}  r r@{\hskip 3mm} r }
    \toprule
    & \multicolumn{2}{c}{PTB} & \multicolumn{2}{c}{CTB}  \\
    Model & Mean  & Max  & Mean  & Max  \\ 
    \midrule
    \hspace{-2mm} PRPN \cite{shen2018nlm} &  37.4 & 38.1& $-$&$-$ \\
    \hspace{-2mm}    ON \cite{shen2019ordered} &  47.7 & 49.4 &$-$&$-$\\
     \hspace{-2mm}   URNNG$^\dag$ \cite{kim2019urnng} & $-$  & 45.4 &$-$&$-$ \\
     \hspace{-2mm}   DIORA$^\dag$ \cite{drozdov2018latent}  \hspace{-3mm} &  $-$ & 58.9 &$-$&$-$\\
         \midrule
   \hspace{-2mm}      Left Branching &  \multicolumn{2}{c}{8.7}  &  \multicolumn{2}{c}{9.7} \\
   \hspace{-2mm}      Right Branching & \multicolumn{2}{c}{39.5} &  \multicolumn{2}{c}{20.0}\\
   \hspace{-2mm}      Random Trees & 19.2 & 19.5 & 15.7 & 16.0\\
   \hspace{-2mm}      PRPN (tuned) & 47.3 & 47.9 & 30.4 & 31.5 \\
   \hspace{-2mm}      ON (tuned) & 48.1 & 50.0 &25.4 & 25.7 \\
   \hspace{-2mm}     Scalar PCFG & \multicolumn{2}{c}{$<\textrm{35.0}$}   & \multicolumn{2}{c}{$<$ 15.0}  \\ 
    \hspace{-2mm}     Neural PCFG & 50.8 & 52.6 & 25.7 & 29.5\\ 
  \hspace{-2mm}       Compound PCFG & 55.2 & 60.1 & 36.0 & 39.8 \\ \midrule
  \hspace{-2mm}       Oracle Trees &  \multicolumn{2}{c}{84.3} & \multicolumn{2}{c}{81.1}  \\
         \bottomrule
    \end{tabular}

    \caption{Unlabeled sentence-level $F_1$ scores on PTB and CTB test sets. Top shows results from previous work while the rest of the results are from this paper. Mean/Max scores are obtained from 4 runs of each model with different random seeds. Oracle is the maximum score obtainable with binarized trees, since we compare against the non-binarized gold trees per convention. Results with $^\dag$ are trained on a version of PTB with punctuation, and hence not strictly comparable to the present work. For URNNG/DIORA, we take the parsed test set provided by the authors from their best runs and evaluate $F_1$ with our evaluation setup, which ignores punctuation.}
    \label{tab:results}
    \end{table}
Given the above, we mainly compare our approach against two recent, strong baselines with open source code: Parsing Predict Reading Network (PRPN)\footnote{\url{https://github.com/yikangshen/PRPN}} \cite{shen2018nlm} and Ordered Neurons (ON)\footnote{\url{https://github.com/yikangshen/Ordered-Neurons}} \cite{shen2019ordered}. These approaches train a neural language model with gated attention-like mechanisms to induce binary trees, and achieve strong unsupervised parsing performance even when trained on corpora where punctuation is removed. Since the original results were on both language modeling and unsupervised parsing, their hyperparameters were presumably tuned to do well on both and thus may not be optimal for just unsupervised parsing. We therefore tune the hyperparameters of these baselines for unsupervised parsing only (i.e. on validation $F_1$).

\section{Results and Discussion}

Table~\ref{tab:results} shows the unlabeled $F_1$ scores for our models and various baselines. All models soundly outperform right branching baselines, and we find that the neural PCFG/compound PCFG are strong models for grammar induction. In particular the compound PCFG outperforms other models by an appreciable margin on both English and Chinese. We again note that we were unable to induce meaningful grammars through a traditional PCFG with the scalar parameterization despite a thorough hyperparameter search.\footnote{Training perplexity was much higher than in the neural case, indicating significant optimization issues. However we did not experiment with online EM \cite{liang2009online}, and it is possible that such methods would yield better results.} See \cref{lab:full} for the full results  broken down by sentence length for sentence- and corpus-level $F_1$.

\begin{table}[]
\small
    \centering
    \begin{tabular}{l c c c c }
    \toprule
    & \multirow{2}{*}{PRPN}& \multirow{2}{*}{ON} & Neural & Compound \\
    &  &  & PCFG &PCFG \\
    \midrule
    Gold & 47.3 & 48.1 & 50.8 & 55.2 \\
    Left & 1.5 & 14.1 & 11.8 & 13.0 \\
    Right & 39.9 & 31.0 & 27.7 & 28.4 \\ 
    Self & 82.3 & 71.3 & 65.2 & 66.8 \\
    \midrule 
    SBAR & 50.0\% & 51.2\% & 52.5\% & 56.1\% \\
    NP & 59.2\% & 64.5\% & 71.2\% & 74.7\% \\
    VP & 46.7\% & 41.0\% & 33.8\% & 41.7\% \\
    PP & 57.2\% & 54.4\% & 58.8\% & 68.8\% \\
    ADJP & 44.3\% & 38.1\% & 32.5\% & 40.4\% \\
    ADVP & 32.8\% & 31.6\% & 45.5\% & 52.5\% \\
         \bottomrule
    \end{tabular}

    \caption{(Top) Mean $F_1$ similarity against Gold, Left, Right, and Self trees. Self $F_1$ score is calculated by averaging over all 6 pairs obtained from 4 different runs. (Bottom) Fraction of ground truth constituents that were predicted as a constituent by the models broken down by label (i.e. label recall).}
    \label{tab:results2}
\end{table}
Table~\ref{tab:results2} analyzes the learned tree structures. We compare similarity  as measured by $F_1$ against gold, left, right, and ``self" trees (top), where self $F_1$ score is calculated by averaging over all 6 pairs obtained from 4 different runs. We find that PRPN is particularly consistent across multiple runs. We also observe that different models are better at identifying different constituent labels, as measured by label recall (Table~\ref{tab:results2}, bottom).
While left as future work, this naturally suggests an ensemble approach wherein the empirical probabilities of constituents (obtained by averaging the predicted binary constituent labels from the different models) are used either to supervise another model or directly as potentials in a CRF constituency  parser. Finally, all models seemed to have some difficulty in identifying SBAR/VP constituents which typically span more words than NP constituents, indicating further opportunities for improvement on unsupervised parsing.
\begin{table}[]
\small
    \centering
    \begin{tabular}{l  r c c }
    \toprule
    & PPL  & Syntactic Eval.  & $F_1$ \\
    \midrule
    LSTM LM & 86.2  &60.9\% & $-$  \\
    PRPN & 87.1 & 62.2\% & 47.9 \\
     \hspace{3mm} Induced RNNG   & 95.3& 60.1\% & 47.8  \\
     \hspace{3mm} Induced URNNG  & 90.1 & 61.8\% & 51.6   \\
    ON & 87.2 & 61.6\% & 50.0\\
     \hspace{3mm} Induced RNNG     & 95.2 & 61.7\% & 50.6  \\
     \hspace{3mm} Induced URNNG  & 89.9 & 61.9\%  & 55.1   \\    
    Neural PCFG &252.6 & 49.2\% & 52.6 \\ 
     \hspace{3mm} Induced RNNG   & 95.8 & 68.1\% & 51.4 \\
     \hspace{3mm} Induced URNNG  & 86.0& 69.1\%  & 58.7   \\    
    Compound PCFG &196.3 & 50.7\%& 60.1\\
     \hspace{3mm} Induced RNNG   &89.8 & 70.0\% & 58.1 \\
     \hspace{3mm} Induced URNNG  & 83.7 & 76.1\% & 66.9   \\    
    \midrule
       RNNG on Oracle Trees   & 80.6 &70.4\% & 71.9  \\
      + URNNG Fine-tuning  & 78.3 & 76.1\% & 72.8 \\   
      \bottomrule
    \end{tabular}
    \caption{Results from training RNNGs on induced trees from various models (Induced RNNG) on the PTB. Induced URNNG indicates fine-tuning with the URNNG objective. We show perplexity (PPL), grammaticality judgment performance (Syntactic Eval.), and unlabeled $F_1$. PPL/$F_1$ are calculated on the PTB test set and Syntactic Eval. is from \citet{marvin2018syntax}'s dataset. Results on top do not make any use of annotated trees, while the bottom two results are trained on binarized gold trees. The perplexity numbers here are not comparable to standard results on the PTB since our models are generative model of sentences and hence we do not carry information across sentence boundaries. Also note that all the RNN-based models above (i.e. LSTM/PRPN/ON/RNNG/URNNG) have roughly the same model capacity (see \cref{lab:rnng}).}
    \label{tab:rnng}
\end{table}

\subsection{Induced Trees for Downstream Tasks}
While the compound PCFG has fewer independence assumptions than the neural PCFG, it is still a more constrained model of language than standard neural language models (NLM) and thus not competitive in terms of perplexity: the compound PCFG obtains a perplexity of 196.3 while an LSTM language model (LM) obtains 86.2 (Table~\ref{tab:rnng}).\footnote{We did manage to almost match the perplexity of an NLM by additionally conditioning the terminal probabilities on previous history, i.e.     \vspace{-4mm}
\begin{align*}
     \pi_{\boldz, T \to w_t} \propto \exp(\boldu^\top_w \,  
f_2([\boldw_T \param \boldz \param \boldh_t]) + b_w),     \\[-6mm]
\end{align*}
where $\boldh_t$ is the hidden state from an LSTM over $\sentence_{<t}$. However the unsupervised parsing performance was far worse ($\approx$ 25 $F_1$ on the PTB).} In contrast, both PRPN and ON perform as well as an LSTM LM while maintaining good unsupervised parsing performance.

\begin{figure}[t]
    \centering
\vspace{-3mm}
    \includegraphics[scale=0.43]{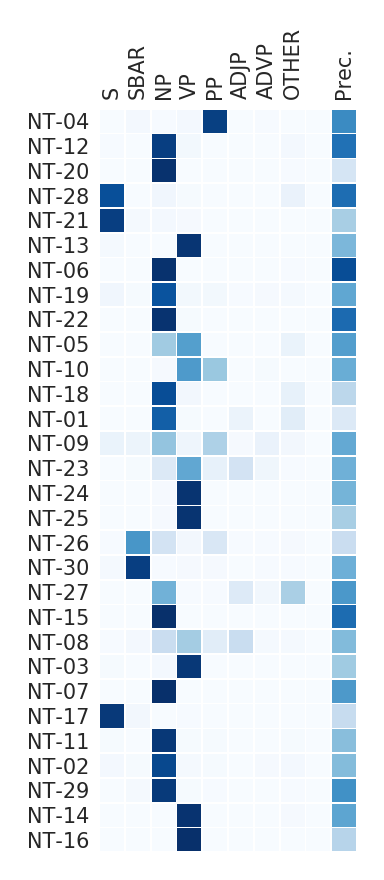}
    \vspace{-2mm}
    \caption{Alignment of induced nonterminals ordered from top based on predicted frequency (therefore NT-04 is the most frequently-predicted nonterminal). For each nonterminal we visualize the proportion of correctly-predicted constituents that correspond to particular gold labels.
    For reference we also show the precision (i.e. probability of correctly predicting unlabeled constituents) in the rightmost column.}
    \label{fig:align}

\end{figure}

We thus experiment to see if it is possible to use the induced trees to supervise a more flexible generative model that can make use of tree structures---namely, recurrent neural network grammars (RNNG) \cite{dyer2016rnng}.
RNNGs are generative models of language that jointly model syntax and surface structure by incrementally generating a syntax tree and sentence. As with NLMs, RNNGs make no  independence assumptions, and have been shown to outperform NLMs in terms of perplexity and grammaticality judgment when trained on gold trees \cite{kuncoro2018syntax,wilcox2019}. 

\begin{table*}[ht]
    \centering
    \tiny
    \begin{tabular}{l}
    \toprule
    \textbf{\textsf{he retired as senior vice president finance and administration and chief financial officer of the company oct. N}} \\
\textsf{kenneth j. $\langle$unk$\rangle$ who was named president of this thrift holding company in august resigned citing personal reasons} \\
\textsf{the former president and chief executive eric w. $\langle$unk$\rangle$ resigned in june} \\
\textsf{$\langle$unk$\rangle$ 's president and chief executive officer john $\langle$unk$\rangle$ said the loss stems from several factors} \\
\textsf{mr. $\langle$unk$\rangle$ is executive vice president and chief financial officer of $\langle$unk$\rangle$ and will continue in those roles} \\
\textsf{charles j. lawson jr. N who had been acting chief executive since june N will continue as chairman} \\
\midrule
\textbf{\textsf{$\langle$unk$\rangle$ corp. received an N million army contract for helicopter engines}} \\
\textsf{boeing co. received a N million air force contract for developing cable systems for the $\langle$unk$\rangle$ missile} \\
\textsf{general dynamics corp. received a N million air force contract for $\langle$unk$\rangle$ training sets} \\
\textsf{grumman corp. received an N million navy contract to upgrade aircraft electronics} \\
\textsf{thomson missile products with about half british aerospace 's annual revenue include the $\langle$unk$\rangle$ $\langle$unk$\rangle$ missile family} \\
\textsf{already british aerospace and french $\langle$unk$\rangle$ $\langle$unk$\rangle$ $\langle$unk$\rangle$ on a british missile contract and on an air-traffic control radar system} \\

\midrule

\textbf{\textsf{meanwhile during the the s\&p trading halt s\&p futures sell orders began $\langle$unk$\rangle$ up while stocks in new york kept falling sharply}} \\
\textsf{but the $\langle$unk$\rangle$ of s\&p futures sell orders weighed on the market and the link with stocks began to fray again} \\
\textsf{on friday some market makers were selling again traders said} \\
\textsf{futures traders say the s\&p was $\langle$unk$\rangle$ that the dow could fall as much as N points} \\
\textsf{meanwhile two initial public offerings $\langle$unk$\rangle$ the $\langle$unk$\rangle$ market in their $\langle$unk$\rangle$ day of national over-the-counter trading friday} \\
\textsf{traders said most of their major institutional investors on the other hand sat tight} \\
  \bottomrule
    \end{tabular}
    \caption{For each query sentence (bold), we show the 5 nearest neighbors based on cosine similarity, where we take the representation for each sentence to be the mean of the variational posterior. }
    \label{tab:nn}
    \vspace{-2mm}
\end{table*}

We take the best run from each model and parse the training set,\footnote{The train/test $F_1$ was similar for all models.} and use the induced trees to supervise an RNNG for each model using the parameterization from \citet{kim2019urnng}.\footnote{\url{https://github.com/harvardnlp/urnng}} We are also interested in syntactic evaluation of our models, and for this we utilize the framework and dataset from \citet{marvin2018syntax}, where a model is presented two minimally different sentences such as: 
\begin{align*}
&\textsf{\small the senators near the assistant are old}  \\ 
&\textsf{*}\textsf{\small the senators near the assistant is old}
\end{align*}
and must assign higher probability to grammatical sentence.

Additionally, \citet{kim2019urnng} report perplexity improvements by fine-tuning an RNNG trained on gold trees with the \emph{unsupervised} RNNG (URNNG)---whereas the RNNG is is trained to maximize the joint log likelihood $\log p(\tree)$, the URNNG maximizes a lower bound on the log marginal likelihood $\log \sum_{\tree \in \mcT_\mcG(\sentence)} p(\tree)$ with a structured inference network that approximates the true posterior. We experiment with a similar approach where we fine-tune RNNGs trained on induced trees with URNNGs. We perform early stopping for both RNNG and URNNG based on validation perplexity. See \cref{lab:rnng} for the full experimental setup. 

The results are shown in Table~\ref{tab:rnng}. For perplexity, RNNGs trained on induced trees (Induced RNNG in Table~\ref{tab:rnng}) are unable to improve upon an LSTM LM, in contrast to the supervised RNNG which does outperform the LSTM language model (Table~\ref{tab:rnng}, bottom). For grammaticality judgment however, the RNNG trained with compound PCFG trees outperforms the LSTM LM despite obtaining worse perplexity,\footnote{\citet{kuncoro2018syntax,kuncoro2019} also observe that models that
achieve lower perplexity do not necessarily perform better on
syntactic evaluation tasks.} and performs on par with the RNNG trained on binarized gold trees.
Fine-tuning with the URNNG results in improvements in perplexity and grammaticality judgment across the board (Induced URNNG in Table~\ref{tab:rnng}). We also obtain large improvements on unsupervised parsing as measured by $F_1$, with the fine-tuned URNNGs  outperforming the respective original models.\footnote{\citet{li2019parsing} similarly obtain improvements by refining a model trained on induced trees on classification tasks.} This is potentially due to an ensembling effect between the original model and the URNNG's structured inference network, which is parameterized as a neural CRF constituency parser \cite{durrett2015crf,liu2018structalign}.\footnote{While left as future work, it is possible to use the compound PCFG itself as an inference network. Also note that the $F_1$ scores for the URNNGs in Table~\ref{tab:rnng} are optimistic since we selected the best-performing runs of the original models based on validation $F_1$ to parse the training set. Finally, as noted by \citet{kim2019urnng}, a URNNG trained from scratch fails to outperform a right-branching baseline on this version of PTB where punctuation is removed.}

\subsection{Model Analysis}\label{sec:analysis}
We analyze our best compound PCFG model
in more detail. Since we induce a full set of nonterminals in our grammar, we can analyze the learned nonterminals to see if they can be aligned with linguistic constituent labels. Figure~\ref{fig:align}  visualizes the alignment between induced and gold labels, where for each nonterminal we show the empirical probability that a predicted constituent of this type will correspond to a particular linguistic constituent in the test set, conditioned on its being a correct constituent (for reference we also show the precision).  We observe that some of the induced nonterminals clearly align to linguistic nonterminals. Further results, including preterminal alignments to part-of-speech tags,\footnote{As a POS induction system, the many-to-one performance of the compound PCFG using the preterminals is 68.0. A similarly-parameterized compound HMM with 60 hidden states (an HMM is a particularly type of PCFG) obtains 63.2. This is still quite a bit lower than the state-of-the-art \cite{tran2016,he2018,stratos2019pos}, though comparison is confounded by various factors such as preprocessing. A neural PCFG/HMM obtains 68.2 and 63.4 respectively.} are shown in \cref{lab:align}.

\begin{table*}[ht]
    \centering
    \tiny
 \begin{tabular}{l}
\toprule 
\hspace{-3mm}
    \tiny
    \parbox{.45\linewidth}{
\centering
    \begin{tabular}{ l@{\hskip 1mm} l }
\multirow{11}{*}{\begin{tikzpicture}
\tiny
\tikzset{level distance=14pt}
\tikzset{sibling distance=-1pt}
\tikzset{edge from parent/.style=
{draw,
edge from parent path={(\tikzparentnode.south)
-- +(0,-3pt)
-| (\tikzchildnode)}}}
\Tree [.NT-04 [.T-13 $w_1$ ] [.NT-12 [.NT-20 [.NT-20 [.NT-07 [.T-05 $w_2$ ] [.T-45 $w_3$ ]] [.T-35 $w_4$ ]] [.T-40 $w_5$ ]] [.T-22 $w_6$ ]]]
\end{tikzpicture}} & 
\textbf{PC -}  \\ 
& \textsf{ of the company 's capital structure}   \\
& \textsf{ in the company 's divestiture program}  \\
& \textsf{ by the company 's new board}   \\
& \textsf{ in the company 's core businesses}  \\
& \textsf{ on the company 's strategic plan} \vspace{1mm}   \\
& \textbf{PC +} \\ 
& \textsf{above the treasury 's N-year note} \\
 &\textsf{above the treasury 's seven-year note} \\
 & \textsf{above the treasury 's comparable note} \\
 &  \textsf{above the treasury 's five-year note} \\
 & \textsf{measured the earth 's ozone layer} \\
    \end{tabular}
    } 
     \hspace{-3mm}

       \parbox{.45\linewidth}{\tiny
\centering
    \begin{tabular}{ l@{\hskip 1mm} l }
\multirow{11}{*}{\begin{tikzpicture}
\tiny
\tikzset{level distance=14pt}
\tikzset{sibling distance=-1pt}
\tikzset{edge from parent/.style=
{draw,
edge from parent path={(\tikzparentnode.south)
-- +(0,-3pt)
-| (\tikzchildnode)}}}
\Tree [.NT-23 [.T-58 $w_1$ ] [.NT-04 [.T-13 $w_2$ ] [.NT-12 [.NT-06 [.T-05 $w_3$ ] [.T-41 $w_4$ ] ] [.NT-04 [.T-13 $w_5$ ] [.NT-12 [.T-60 $w_6$ ] [.T-21 $w_7$ ]]]]]]
\end{tikzpicture}} & 
\textbf{PC -}  \\ 
& \textsf{ purchased through the exercise of stock options} \\
& \textsf{ circulated by a handful of major brokers}  \\
& \textsf{ higher as a percentage of total loans}  \\
& \textsf{ common with a lot of large companies}  \\
& \textsf{ surprised by the storm of sell orders} \vspace{1mm}  \\
& \textbf{PC +}  \\ 
 & \textsf{brought to the u.s. against her will} \\
& \textsf{laid for the arrest of opposition activists} \\
 & \textsf{uncertain about the magnitude of structural damage} \\
 & \textsf{held after the assassination of his mother} \\
 & \textsf{hurt as a result of the violations} 
    \end{tabular}}
\vspace{1mm} \\
\line(1,0){440} \\
\hspace{-3mm}
         \parbox{.45\linewidth}{\tiny
\centering

    \begin{tabular}{ l@{\hskip 1mm} l }
\multirow{11}{*}{\begin{tikzpicture}
\tiny
\tikzset{level distance=14pt}
\tikzset{sibling distance=-1pt}
\tikzset{edge from parent/.style=
{draw,
edge from parent path={(\tikzparentnode.south)
-- +(0,-3pt)
-| (\tikzchildnode)}}}
\Tree [.NT-10 [.T-55 $w_1$ ] [.NT-05 [.T-02 $w_2$ ] [.NT-19 [.NT-06 [.T-05 $w_3$ ] [.T-41 $w_4$ ] ] [.NT-04 [.T-13 $w_5$ ] [.T-43 $w_6$ ] ] ] ] ]
\end{tikzpicture}} & 
\textbf{PC -}  \\ 
& \textsf{ to terminate their contract with warner}  \\
& \textsf{ to support a coup in panama} \\
& \textsf{ to suit the bureaucrats in brussels}  \\
& \textsf{ to thwart his bid for amr} \\
&\textsf{ to prevent the pound from rising} \vspace{1mm}   \\
& \textbf{PC +}  \\ 
& \textsf{to change our strategy of investing} \\
& \textsf{to offset the growth of minimills} \\
& \textsf{to be a lot of art} \\
 & \textsf{to change our way of life} \\
& \textsf{to increase the impact of advertising} 
    \end{tabular}}
    \hspace{-3mm}
    \parbox{.45\linewidth}{
\centering
   \begin{tabular}{ l@{\hskip 1mm} l }
    
\multirow{11}{*}{\begin{tikzpicture}
\tiny
\tikzset{level distance=14pt}
\tikzset{sibling distance=-1pt}
\tikzset{edge from parent/.style=
{draw,
edge from parent path={(\tikzparentnode.south)
-- +(0,-3pt)
-| (\tikzchildnode)}}}
\Tree [.NT-05 [.T-02 $w_1$ ] [.NT-19 [.NT-06 [.NT-20 [.T-05 $w_2$ ] [.T-40 $w_3$ ]] [.T-22 $w_4$ ]] [.NT-04 [.T-13 $w_5$ ] [.NT-12 [.T-60 $w_6$ ] [.T-21 $w_7$ ]]]]]
\end{tikzpicture}} & 
\textbf{PC -} \\ 
& \textsf{ raise the minimum grant for smaller states}  \\
&\textsf{ veto a defense bill with inadequate funding}  \\
& \textsf{ avoid an imminent public or private injury} \\
& \textsf{ field a competitive slate of congressional candidates}  \\
& \textsf{ alter a longstanding ban on such involvement}  \vspace{1mm}   \\
& \textbf{PC +} \\ 
& \textsf{generate an offsetting profit by selling waves} \\
& \textsf{change an export loss to domestic plus} \\
 & \textsf{expect any immediate problems with margin calls} \\
 & \textsf{make a positive contribution to our earnings} \\
 & \textsf{find a trading focus discouraging much participation} \\
    \end{tabular}}
\\ \bottomrule
\end{tabular}

    \caption{For each subtree, we perform PCA on the variational posterior mean vectors that are associated with that particular subtree and take the top principal component. We then list the top 5 constituents that had the lowest (\textbf{PC -}) and highest (\textbf{PC +}) principal component values.}
    \label{tab:pca2}
    \vspace{-2mm}
\end{table*}

We next analyze the continuous latent space. Table~\ref{tab:nn} shows 
nearest neighbors of some sentences using the mean of the variational posterior as the continuous representation of each sentence. We qualitatively observe that the latent space seems to capture topical information. 

We are also interested in the variation in the leaves due to $\boldz$ when the variation due to the tree structure is held constant. To investigate this, we use the parsed dataset to obtain pairs of the form $(\boldsymbol{\mu}_\phi(\sentence^{(n)}), \tree^{(n)}_j)$, where $\tree^{(n)}_j$ is the $j$-th subtree of the (approximate) MAP tree $\tree^{(n)}$ for the $n$-th sentence. Therefore each mean vector $\boldsymbol{\mu}_\phi(\sentence^{(n)})$ is associated with $|\sentence^{(n)}| -1$ subtrees, where $|\sentence^{(n)}|$ is the sentence length. Our definition of subtree here ignores terminals, and thus each subtree is associated with many mean vectors. For a frequently occurring subtree, we perform PCA on the set of mean vectors that are associated with the subtree to obtain the top principal component. We then show the constituents that had the 5 most positive/negative values for this top principal component in Table~\ref{tab:pca2}. For example, a particularly common subtree---associated with 180 unique constituents---is given by 
\begin{align*}
    &\text{\small (NT-04 (T-13 $w_1$) (NT-12 (NT-20 (NT-20 (NT-07 (T-05 $w_2$) } \\
    &\hspace*{4mm} \text{\small (T-45 $w_3$)) (T-35 $w_4$)) (T-40 $w_5$)) (T-22 $w_6$)))}.
\end{align*}
The top 5 constituents with the most negative/positive values are shown in the top left part of Table~\ref{tab:pca2}. We find that the leaves $[w_1, \dots, w_6]$, which form a 6-word constituent, vary in a regular manner as $\boldz$ is varied. We also observe that root of this subtree (NT-04) aligns to prepositional phrases (PP) in Figure~\ref{fig:align}, and the leaves in Table~\ref{tab:pca2} (top left) are indeed mostly PP. However, the model fails to identify {\small ((T-40 $w_5$) (T-22 $w_6$))} as a constituent in this case (as well as well in the bottom right example). See appendix~\ref{lab:subtrees} for more examples.
It is possible that  the model is utilizing the subtrees to capture broad template-like structures and then using $\boldz$ to fill them in, similar to recent works that also train models to separate ``what to say" from ``how to say it" \cite{wiseman2018templates,peng2019exemplar,chen2019exemplar,chen2019multi}.

\subsection{Limitations} 
We report on some negative results as well as important limitations of our work.
While distributed representations promote parameter sharing, we were unable to obtain improvements through more factorized parameterizations that promote even greater parameter sharing. In particular, for rules of the type $A \to BC$, we tried  having the output embeddings be a function of the input embeddings (e.g. $\boldu_{BC} = g([\boldw_B \param \boldw_C])$ where $g$ is an MLP), but obtained worse results. For rules of the type $T \to w$, we tried using a character-level CNN \cite{santos2014char,kim2016char} to obtain the output word embeddings $\boldu_w$ \cite{joze2016lm,tran2016}, but found the performance to be similar to the word-level case.\footnote{It is also possible to take advantage of pretrained word embeddings by using them to initialize output word embeddings or directly working with continuous emission distributions \cite{lin2015pos,he2018}} We were also unable to obtain improvements by making the variational family more flexible through normalizing flows \cite{rezende2015flow,kingma2016flow}. However, given that we did not exhaustively explore the full space of possible parameterizations, the above modifications could eventually lead to improvements with the right setup. 

Relatedly, the models were quite sensitive to parameterization (e.g. it was important to use residual layers for $f_1, f_2$), grammar size, and optimization method. We also noticed some variance in results across random seeds, as shown in Table~\ref{tab:results2}.
Finally, despite vectorized GPU implementations, training was significantly more expensive (both in terms of time and memory) than NLM-based unsupervised parsing systems due to the $O(|\mcR||\sentence|^3)$ dynamic program, which makes our approach potentially difficult to scale.

\section{Related Work}

Grammar induction and unsupervised parsing has a  long and rich history in natural language processing. Early work on with pure unsupervised learning was mostly negative \cite{lari1990scfg,carroll1992two,charniak1993}, though \citet{pereira1992io} reported some success on partially bracketed data. \citet{clark2001pcfg} and \citet{klein2002ccm}
were some of the first successful statistical approaches. In particular, the constituent-context model (CCM) of \citet{klein2002ccm}, which explicitly models both constituents and distituents, was the basis for much subsequent work \cite{klein2004syntax,huang2012,golland2012}.
Other works have explored imposing inductive biases through Bayesian  priors \cite{johnson2007pcfg,liang2007pcfg,wang2013collapsed}, modified objectives \cite{smith2004}, and additional constraints on recursion depth \cite{noji2016left,jin2018depth}.

While the framework of specifying the structure of a grammar and learning the parameters is common, other methods exist. \citet{bod2006subtrees} consider a nonparametric-style approach to unsupervised parsing by using random subsets of training subtrees to parse new sentences. \citet{seginer2007unsup} utilize an incremental algorithm to unsupervised parsing which makes local decisions to create constituents based on a complex set of heuristics. \citet{ponvert2011simple} induce parse trees through cascaded applications of finite state models.

More recently, neural network-based approaches have shown promising results on inducing parse trees directly from words.
\citet{shen2018nlm,shen2019ordered} learn tree structures through soft gating layers within neural language models, while \citet{drozdov2018latent} combine recursive autoencoders with the inside-outside algorithm. \citet{kim2019urnng} train unsupervised recurrent neural network grammars with a structured inference network to induce latent trees, and \citet{shi2019visual} utilize image captions to identify and ground constituents.

Our work is also related to latent variable PCFGs \cite{matsu2005pcfg,petrov2006lpcfg,cohen2012spec}, which extend PCFGs to the latent variable setting by splitting nonterminal symbols into latent subsymbols. In particular, latent vector grammars \cite{zhao2018latent} and compositional vector grammars \cite{socher2013cvg} also employ continuous vectors within their grammars. However these approaches have been employed for learning supervised parsers on annotated treebanks, in contrast to the unsupervised setting of the current work.

\section{Conclusion}

This work studies a neural network-based approach grammar induction with PCFGs. We first propose to parameterize a PCFG's rule probabilities with neural networks over distributed representations of latent symbols, and find that this neural PCFG makes it possible to induce linguistically meaningful grammars with simple maximum likelihood learning. We then extend the neural PCFG through a sentence-level continuous latent vector, which induces marginal dependencies beyond the traditional first-order context-free assumptions. We show that this compound PCFG learns richer grammars and leads to improved performance when evaluated as an unsupervised parser.
The collapsed amortized variational inference approach is general and can be used for generative models which admit tractable inference through partial conditioning. Learning deep generative models which exhibit such conditional Markov properties is an interesting direction for future work.

\section*{Acknowledgments}

We thank Phil Blunsom for initial discussions which seeded many of the core ideas in the present work. We also thank Yonatan Belinkov and Shay Cohen for helpful feedback, and Andrew Drozdov for providing the parsed dataset from their DIORA model. YK is supported by a Google Fellowship. AMR acknowledges the support of NSF 1704834, 1845664, AWS, and Oracle.

{\footnotesize
\bibliography{master}
\bibliographystyle{acl_natbib}}

\appendix

\newpage
\label{sec:appendix}
\section{Appendix}

\subsection{Model Parameterization}\label{lab:model}
We associate an input embedding $\boldw_N$ for each symbol $N$ on the left side of a rule (i.e. $N \in \{S\} \cup  \mcN \cup \mcP$) and run a neural network over $\boldw_N$ to obtain the rule probabilities. Concretely, each rule type $\pi_r$ is parameterized as follows,
\begin{align*}
   \pi_{S \to A} &= \frac{\exp(\boldu^\top_A \,  f_1(\boldw_S)  + b_A)}{\sum_{A' \in \mcN}\exp(\boldu^\top_{A'} \,  f_1(\boldw_S) + b_{A'} ) },  \\
    \pi_{A \to BC} &= \frac{\exp(\boldu^\top_{BC} \, \boldw_{A} + b_{BC})}{\sum_{B'C' \in \mcM} \exp(\boldu^\top_{B'C'} \, \boldw_{A} + b_{B'C'})}, \\ 
    \pi_{T \to w} &= \frac{\exp(\boldu^\top_w \,  f_2(\boldw_T) + b_w)}{\sum_{w' \in \Sigma}\exp(\boldu^\top_{w'} \,  f_2(\boldw_T)  + b_{w'}) } , \\
\end{align*}
where $\mcM$ is the product space $(\mcN \cup \mcP) \times (\mcN \cup \mcP)$, and $f_1, f_2$ are MLPs with two residual layers,
\begin{align*}
    f_i(\boldx) = &g_{i, 1}(g_{i, 2}(\mathbf{W}_i \boldx + \boldb_i)), \\
    g_{i,j}(\boldy) =& \relu(\mathbf{V}_{i,j}\relu(\mathbf{U}_{i,j}\boldy + \boldp_{i,j}) + \\& \boldq_{i,j}) + \boldy.
\end{align*}

In the compound PCFG the rule probabilities $\bpi_\boldz$ given a latent vector $\boldz$,
\begin{align*}
   \pi_{\boldz, S \to A}^{} & = \frac{\exp(\boldu^\top_A \,  f_1([\boldw_S \param \boldz^{}]) + b_A)}{\sum_{A' \in \mcN}\exp(\boldu^\top_{A'} \,  f_1([\boldw_S; \boldz^{}]) + b_{A'}) },\\
    \pi_{\boldz, A \to B C}^{} & = \frac{\exp(\boldu^\top_{BC} \, [\boldw_{A} \param \boldz^{}]  + b_{BC})}{\sum_{B'C' \in \mcM} \exp(\boldu^\top_{B'C'} \, [\boldw_{A} \param \boldz^{}] + b_{B'C'})}, \\ 
    \pi_{\boldz, T \to w}^{} &=  \frac{\exp(\boldu^\top_w \,  f_2([\boldw_T \param \boldz^{}]) + b_{w})}{\sum_{w' \in \Sigma}\exp(\boldu^\top_{w'} \,  f_2([\boldw_T \param \boldz^{}]) + b_{w'}) }.
\end{align*}
Again $f_1, f_2$ are as before where the first layer's input dimensions are appropriately changed to account for concatenation with $\boldz^{}$. 

\begin{table*}[h]
\small
    \centering
    \begin{tabular}{l r  r r r r}
    \toprule
    & \multicolumn{5}{c}{Sentence-level $F_1$}   \\
   &  WSJ-10 & WSJ-20  & WSJ-30 & WSJ-40 & WSJ-Full   \\ 
    \midrule
         Left Branching & 17.4 & 12.9 & 9.9 & 8.6 & 8.7 \\
         Right Branching & 58.5 & 49.8 & 44.4 & 41.6 & 39.5 \\
         Random Trees & 31.8 & 25.2 & 21.5 & 19.7 & 19.2 \\
         PRPN (tuned) & 58.4 & 54.3 & 50.9 & 48.5  & 47.3\\
         ON (tuned)  & 63.9 & 57.5 & 53.2 & 50.5 &  48.1 \\
         Neural PCFG & 64.6 & 58.1 & 54.6 & 52.6 & 50.8 \\ 
         Compound PCFG & 70.5 & 63.4 & 58.9 & 56.6 & 55.2 \\
         \midrule
         Oracle & 82.1 & 84.1 & 84.2 & 84.3 & 84.3 \\
\bottomrule  \\
\toprule 
    & \multicolumn{5}{c}{Corpus-level $F_1$}   \\
   &  WSJ-10 & WSJ-20  & WSJ-30 & WSJ-40 & WSJ-Full   \\ 
    \midrule
         Left Branching & 16.5 & 11.7 & 8.5 & 7.2 & 6.0 \\
         Right Branching & 58.9 & 48.3 & 42.5 & 39.4 & 36.1 \\
         Random Trees & 31.9 & 23.9 & 20.0 & 18.1 & 16.4 \\
         PRPN (tuned) & 59.3 &  53.6 & 49.7 & 46.9 & 44.5   \\
         ON (tuned)  & 64.7 & 56.3 & 51.5 & 48.3 &  45.6 \\
         Neural PCFG & 63.5 & 56.8 & 53.1 & 51.0 & 48.7 \\ 
         Compound PCFG & 70.6 & 62.0 & 57.1 & 54.6 & 52.4 \\
         \midrule
         Oracle & 83.5 & 85.2 & 84.9 & 84.9 & 84.7 \\
         \bottomrule
    \end{tabular}    
    \caption{Average unlabeled $F_1$ for the various models broken down by sentence length on the PTB test set. For example WSJ-10 refers to $F_1$ calculated on the subset of the test set where the maximum sentence length is at most 10. Scores are averaged across 4 runs of the model with different random seeds. Oracle is the performance of binarized gold trees (with right branching binarization). Top shows  sentence-level $F_1$ and bottom shows corpus-level $F_1$.} 
    \label{tab:results3}

\end{table*}
\subsection{Corpus/Sentence $F_1$ by Sentence Length}\label{lab:full} For completeness we show the corpus-level and sentence-level $F_1$ broken down by sentence length in Table~\ref{tab:results3}, averaged across 4 different runs of each model. 
In Figure~\ref{fig:gm}  we use the following to refer to rule probabilities of different rule types for the neural PCFG (left),
\begin{align*}
    \pi_S &= \{\pi_r \,\vert \,r \in L(S)\}, \\
    \pi_{\mcN} &= \{\pi_r \,\vert\, r \in L(A), A \in \mcN \} ,\\
    \pi_{\mcP} &= \{\pi_r \,\vert\, r \in L(T), T \in \mcP \}, \\
    \bpi &= \pi_S \cup \pi_{\mcN} \cup \pi_{\mcP},
\end{align*}
where $L(A)$ denotes the set of rules with $A$ on the left hand side. The set of rule probabilities for the compound PCFG (right) is similarly defined,
\begin{align*}
    \pi_{\boldz, S} &= \{\pi_{\boldz, r} \,\vert \,r \in L(S)\}, \\
    \pi_{\boldz, \mcN} &= \{\pi_{\boldz, r} \,\vert\, r \in L(A), A \in \mcN \} ,\\
    \pi_{\boldz, \mcP} &= \{\pi_{\boldz, r} \,\vert\, r \in L(T), T \in \mcP \}, \\
    \bpi_{\boldz} &=  \pi_{\boldz, S}\cup \pi_{\boldz, \mcN} \cup \pi_{\boldz, \mcP}.
\end{align*}

\begin{table*}[h]
    \centering
    \small
    \begin{tabular}{r r r r r r r r  r r r r r }
    \toprule
         Label & S& SBAR & NP & VP & PP & ADJP & ADVP & Other &&& Freq. & Acc.   \\
         \midrule 
NT-01 & 0.0\% & 0.0\% & 81.8\% & 1.1\% & 0.0\% & 5.9\% & 0.0\% & 11.2\% & & &2.9\% & 13.8\%\\
NT-02 & 2.2\% & 0.9\% & 90.8\% & 1.7\% & 0.9\% & 0.0\% & 1.3\% & 2.2\% & & &1.1\% & 44.0\%\\
NT-03 & 1.0\% & 0.0\% & 2.3\% & 96.8\% & 0.0\% & 0.0\% & 0.0\% & 0.0\% & & &1.8\% & 37.1\%\\
NT-04 & 0.3\% & 2.2\% & 0.5\% & 2.0\% & 93.9\% & 0.2\% & 0.6\% & 0.3\% & & &11.0\% & 64.9\%\\
NT-05 & 0.2\% & 0.0\% & 36.4\% & 56.9\% & 0.0\% & 0.0\% & 0.2\% & 6.2\% & & &3.1\% & 57.1\%\\
NT-06 & 0.0\% & 0.0\% & 99.1\% & 0.0\% & 0.1\% & 0.0\% & 0.2\% & 0.6\% & & &5.2\% & 89.0\%\\
NT-07 & 0.0\% & 0.0\% & 99.7\% & 0.0\% & 0.3\% & 0.0\% & 0.0\% & 0.0\% & & &1.3\% & 59.3\%\\
NT-08 & 0.5\% & 2.2\% & 23.3\% & 35.6\% & 11.3\% & 23.6\% & 1.7\% & 1.7\% & & &2.0\% & 44.3\%\\
NT-09 & 6.3\% & 5.6\% & 40.2\% & 4.3\% & 32.6\% & 1.2\% & 7.0\% & 2.8\% & & &2.6\% & 52.1\%\\
NT-10 & 0.1\% & 0.1\% & 1.4\% & 58.8\% & 38.6\% & 0.0\% & 0.8\% & 0.1\% & & &3.0\% & 50.5\%\\
NT-11 & 0.9\% & 0.0\% & 96.5\% & 0.9\% & 0.9\% & 0.0\% & 0.0\% & 0.9\% & & &1.1\% & 42.9\%\\
NT-12 & 0.5\% & 0.2\% & 94.4\% & 2.4\% & 0.2\% & 0.1\% & 0.2\% & 2.0\% & & &8.9\% & 74.9\%\\
NT-13 & 1.6\% & 0.1\% & 0.2\% & 97.7\% & 0.2\% & 0.1\% & 0.1\% & 0.1\% & & &6.2\% & 46.0\%\\
NT-14 & 0.0\% & 0.0\% & 0.0\% & 98.6\% & 0.0\% & 0.0\% & 0.0\% & 1.4\% & & &0.9\% & 54.1\%\\
NT-15 & 0.0\% & 0.0\% & 99.7\% & 0.0\% & 0.3\% & 0.0\% & 0.0\% & 0.0\% & & &2.0\% & 76.9\%\\
NT-16 & 0.0\% & 0.0\% & 0.0\% & 100.0\% & 0.0\% & 0.0\% & 0.0\% & 0.0\% & & &0.3\% & 29.9\%\\
NT-17 & 96.4\% & 2.9\% & 0.0\% & 0.7\% & 0.0\% & 0.0\% & 0.0\% & 0.0\% & & &1.2\% & 24.4\%\\
NT-18 & 0.3\% & 0.0\% & 88.7\% & 2.8\% & 0.3\% & 0.0\% & 0.0\% & 7.9\% & & &3.0\% & 28.3\%\\
NT-19 & 3.9\% & 1.0\% & 86.6\% & 2.4\% & 2.6\% & 0.4\% & 1.3\% & 1.8\% & & &4.5\% & 53.4\%\\
NT-20 & 0.0\% & 0.0\% & 99.0\% & 0.0\% & 0.0\% & 0.3\% & 0.2\% & 0.5\% & & &7.4\% & 17.5\%\\
NT-21 & 94.4\% & 1.7\% & 2.0\% & 1.4\% & 0.3\% & 0.1\% & 0.0\% & 0.1\% & & &6.2\% & 34.7\%\\
NT-22 & 0.1\% & 0.0\% & 98.4\% & 1.1\% & 0.1\% & 0.0\% & 0.2\% & 0.2\% & & &3.5\% & 77.6\%\\
NT-23 & 0.4\% & 0.9\% & 14.0\% & 53.1\% & 8.2\% & 18.5\% & 4.3\% & 0.7\% & & &2.4\% & 49.1\%\\
NT-24 & 0.0\% & 0.2\% & 1.5\% & 98.3\% & 0.0\% & 0.0\% & 0.0\% & 0.0\% & & &2.3\% & 47.3\%\\
NT-25 & 0.3\% & 0.0\% & 1.4\% & 98.3\% & 0.0\% & 0.0\% & 0.0\% & 0.0\% & & &2.2\% & 34.6\%\\
NT-26 & 0.4\% & 60.7\% & 18.4\% & 3.0\% & 15.4\% & 0.4\% & 0.4\% & 1.3\% & & &2.1\% & 23.4\%\\
NT-27 & 0.0\% & 0.0\% & 48.7\% & 0.5\% & 0.7\% & 13.1\% & 3.2\% & 33.8\% & & &2.0\% & 59.7\%\\
NT-28 & 88.2\% & 0.3\% & 3.8\% & 0.9\% & 0.1\% & 0.0\% & 0.0\% & 6.9\% & & &6.7\% & 76.5\%\\
NT-29 & 0.0\% & 1.7\% & 95.8\% & 1.0\% & 0.7\% & 0.0\% & 0.0\% & 0.7\% & & &1.0\% & 62.8\%\\
NT-30 & 1.6\% & 94.5\% & 0.6\% & 1.2\% & 1.2\% & 0.0\% & 0.4\% & 0.4\% & & &2.1\% & 49.4\%\\
\midrule
         \midrule 
NT-01 & 0.0\% & 0.0\% & 0.0\% & 99.2\% & 0.0\% & 0.0\% & 0.0\% & 0.8\% & & &2.6\% & 41.1\%\\
NT-02 & 0.0\% & 0.3\% & 0.3\% & 99.2\% & 0.0\% & 0.0\% & 0.0\% & 0.3\% & & &5.3\% & 15.4\%\\
NT-03 & 88.2\% & 0.3\% & 3.6\% & 1.0\% & 0.1\% & 0.0\% & 0.0\% & 6.9\% & & &7.2\% & 71.4\%\\
NT-04 & 0.0\% & 0.0\% & 100.0\% & 0.0\% & 0.0\% & 0.0\% & 0.0\% & 0.0\% & & &0.5\% & 2.4\%\\
NT-05 & 0.0\% & 0.0\% & 0.0\% & 96.6\% & 0.0\% & 0.0\% & 0.0\% & 3.4\% & & &5.0\% & 1.2\%\\
NT-06 & 0.0\% & 0.4\% & 0.4\% & 98.8\% & 0.0\% & 0.0\% & 0.0\% & 0.4\% & & &1.2\% & 43.7\%\\
NT-07 & 0.2\% & 0.0\% & 95.3\% & 0.9\% & 0.0\% & 1.6\% & 0.1\% & 1.9\% & & &2.8\% & 60.6\%\\
NT-08 & 1.0\% & 0.4\% & 95.3\% & 2.3\% & 0.4\% & 0.2\% & 0.3\% & 0.2\% & & &9.4\% & 63.0\%\\
NT-09 & 0.6\% & 0.0\% & 87.4\% & 1.9\% & 0.0\% & 0.0\% & 0.0\% & 10.1\% & & &1.0\% & 33.8\%\\
NT-10 & 78.3\% & 17.9\% & 3.0\% & 0.5\% & 0.0\% & 0.0\% & 0.0\% & 0.3\% & & &1.9\% & 42.0\%\\
NT-11 & 0.3\% & 0.0\% & 99.0\% & 0.3\% & 0.0\% & 0.3\% & 0.0\% & 0.0\% & & &0.9\% & 70.3\%\\
NT-12 & 0.0\% & 8.8\% & 76.5\% & 2.9\% & 5.9\% & 0.0\% & 0.0\% & 5.9\% & & &2.0\% & 3.6\%\\
NT-13 & 0.5\% & 2.0\% & 1.0\% & 96.6\% & 0.0\% & 0.0\% & 0.0\% & 0.0\% & & &1.7\% & 50.7\%\\
NT-14 & 0.0\% & 0.0\% & 99.1\% & 0.0\% & 0.0\% & 0.6\% & 0.0\% & 0.4\% & & &7.7\% & 14.8\%\\
NT-15 & 2.9\% & 0.5\% & 0.4\% & 95.5\% & 0.4\% & 0.0\% & 0.0\% & 0.2\% & & &4.4\% & 45.2\%\\
NT-16 & 0.4\% & 0.4\% & 17.9\% & 5.6\% & 64.1\% & 0.4\% & 6.8\% & 4.4\% & & &1.4\% & 38.1\%\\
NT-17 & 0.1\% & 0.0\% & 98.2\% & 0.5\% & 0.1\% & 0.1\% & 0.1\% & 0.9\% & & &9.6\% & 85.4\%\\
NT-18 & 0.1\% & 0.0\% & 95.7\% & 1.6\% & 0.0\% & 0.1\% & 0.2\% & 2.3\% & & &4.7\% & 56.2\%\\
NT-19 & 0.0\% & 0.0\% & 98.9\% & 0.0\% & 0.4\% & 0.0\% & 0.0\% & 0.7\% & & &1.3\% & 72.6\%\\
NT-20 & 2.0\% & 22.7\% & 3.0\% & 4.8\% & 63.9\% & 0.6\% & 2.3\% & 0.6\% & & &6.8\% & 59.0\%\\
NT-21 & 0.0\% & 0.0\% & 14.3\% & 42.9\% & 0.0\% & 0.0\% & 42.9\% & 0.0\% & & &2.2\% & 0.7\%\\
NT-22 & 1.4\% & 0.0\% & 11.0\% & 86.3\% & 0.0\% & 0.0\% & 0.0\% & 1.4\% & & &1.0\% & 15.2\%\\
NT-23 & 0.1\% & 0.0\% & 58.3\% & 0.8\% & 0.4\% & 5.0\% & 1.7\% & 33.7\% & & &2.8\% & 62.7\%\\
NT-24 & 0.0\% & 0.0\% & 100.0\% & 0.0\% & 0.0\% & 0.0\% & 0.0\% & 0.0\% & & &0.6\% & 70.2\%\\
NT-25 & 2.2\% & 0.0\% & 76.1\% & 4.3\% & 0.0\% & 2.2\% & 0.0\% & 15.2\% & & &0.4\% & 23.5\%\\
NT-26 & 0.0\% & 0.0\% & 2.3\% & 94.2\% & 3.5\% & 0.0\% & 0.0\% & 0.0\% & & &0.8\% & 24.0\%\\
NT-27 & 96.6\% & 0.2\% & 1.5\% & 1.1\% & 0.3\% & 0.2\% & 0.0\% & 0.2\% & & &4.3\% & 32.2\%\\
NT-28 & 1.2\% & 3.7\% & 1.5\% & 5.8\% & 85.7\% & 0.9\% & 0.9\% & 0.3\% & & &7.6\% & 64.9\%\\
NT-29 & 3.0\% & 82.0\% & 1.5\% & 13.5\% & 0.0\% & 0.0\% & 0.0\% & 0.0\% & & &0.6\% & 45.4\%\\
NT-30 & 0.0\% & 0.0\% & 1.0\% & 60.2\% & 19.4\% & 1.9\% & 4.9\% & 12.6\% & & &2.1\% & 10.4\%\\
\midrule
Gold   &  15.0\% & 4.8\% & 38.5\% & 21.7\% & 14.6\% & 1.7\% & 0.8\% & 2.9\% \\
         \bottomrule
    \end{tabular}
    \vspace{-2mm}
    \caption{Analysis of label alignment for nonterminals in the compound PCFG (top) and the neural PCFG (bottom). Label alignment is the proportion of correctly-predicted constistuents that correspond to a particular gold label. We also show the predicted constituent frequency and accuracy (i.e. precision) on the right. Bottom line shows the frequency in the gold trees.}
    \vspace{-2mm}
    \label{tab:labels}
\end{table*}

\subsection{Experiments with RNNGs}\label{lab:rnng} For experiments on supervising RNNGs with induced trees, we use the parameterization and hyperparameters from \citet{kim2019urnng}, which uses a 2-layer 650-dimensional stack LSTM (with dropout of 0.5) and a 650-dimensional tree LSTM \cite{tai2015treelstm,zhu2015treelstm} as the composition function.

Concretely, the generative story is as follows: first, the stack representation is used to predict the next action (\textsc{shift} or \textsc{reduce}) via an affine transformation followed by a sigmoid. If \textsc{shift} is chosen, we obtain a distribution over the vocabulary via another affine transformation over the stack representation followed by a softmax. Then we sample the next word from this distribution and shift the generated word onto the stack using the stack LSTM. If \textsc{reduce} is chosen, we pop the last two elements off the stack and use the tree LSTM to obtain a new representation. This new representation is shifted onto the stack via the stack LSTM. Note that this RNNG parameterization is slightly different than the original from \citet{dyer2016rnng}, which does not ignore constituent labels and utilizes a bidirectional LSTM as the composition function instead of a tree LSTM. As our RNNG parameterization only works with binary trees, we binarize the gold trees with right binarization for the RNNG trained on gold trees (trees from the unsupervised methods explored in this paper are already binary). 
The RNNG also trains a discriminative parser alongside the generative model for evaluation with importance sampling. We use a CRF parser whose span score parameterization is similar similar to recent works \cite{wang2016graph,stern2017minimal,kitaev2018parsing}: position embeddings are added to word embeddings, and a bidirectional LSTM with 256 hidden dimensions is run over the input representations to obtain the forward  and backward hidden states. The score $s_{ij}\in \reals$ for a constituent spanning the $i$-th and $j$-th word is given by,
\begin{equation*}
 s_{ij} = \MLP([\overrightarrow{\boldh}_{j+1} - \overrightarrow{\boldh}_{i}; \overleftarrow{\boldh}_{i-1} - \overleftarrow{\boldh}_{j}]), 
\end{equation*}
 where the MLP  has a single hidden layer with $\relu$ nonlinearity followed by layer normalization \cite{ba2016layernorm}. 
 
 For experiments on fine-tuning the RNNG with the unsupervised RNNG, we take the discriminative parser (which is also pretrained alongside the RNNG on induced trees) to be the structured inference network for optimizing the evidence lower bound. We refer the reader to \citet{kim2019urnng} and their open source implementation\footnote{\url{https://github.com/harvardnlp/urnng}} for additional details. We also observe that as noted by \citet{kim2019urnng}, a URNNG trained from scratch on this version of PTB without punctuation failed to outperform a right-branching baseline.
 
The LSTM language model baseline is the same size as the stack LSTM (i.e. 2 layers, 650 hidden units, dropout of 0.5), and is therefore equivalent to an RNNG with completely right branching trees. The PRPN/ON baselines for perplexity/syntactic evaluation in Table~\ref{tab:rnng} also have 2 layers with 650 hidden units and 0.5 dropout. Therefore all models considered in Table~\ref{tab:rnng} have roughly the same capacity. For all models we share input/output word embeddings \cite{press2016tie}. Perplexity estimation for the RNNGs and the compound PCFG uses 1000 importance-weighted samples.    

For grammaticality judgment, we modify the publicly available dataset from 
\citet{marvin2018syntax}\footnote{\url{https://github.com/BeckyMarvin/LM_syneval}} to only keep sentence pairs that did not have any unknown words with respect to our PTB vocabulary of 10K words. This results in 33K sentence pairs for evaluation. 
\begin{figure*}
    \centering
    \small
    \includegraphics[scale=0.5]{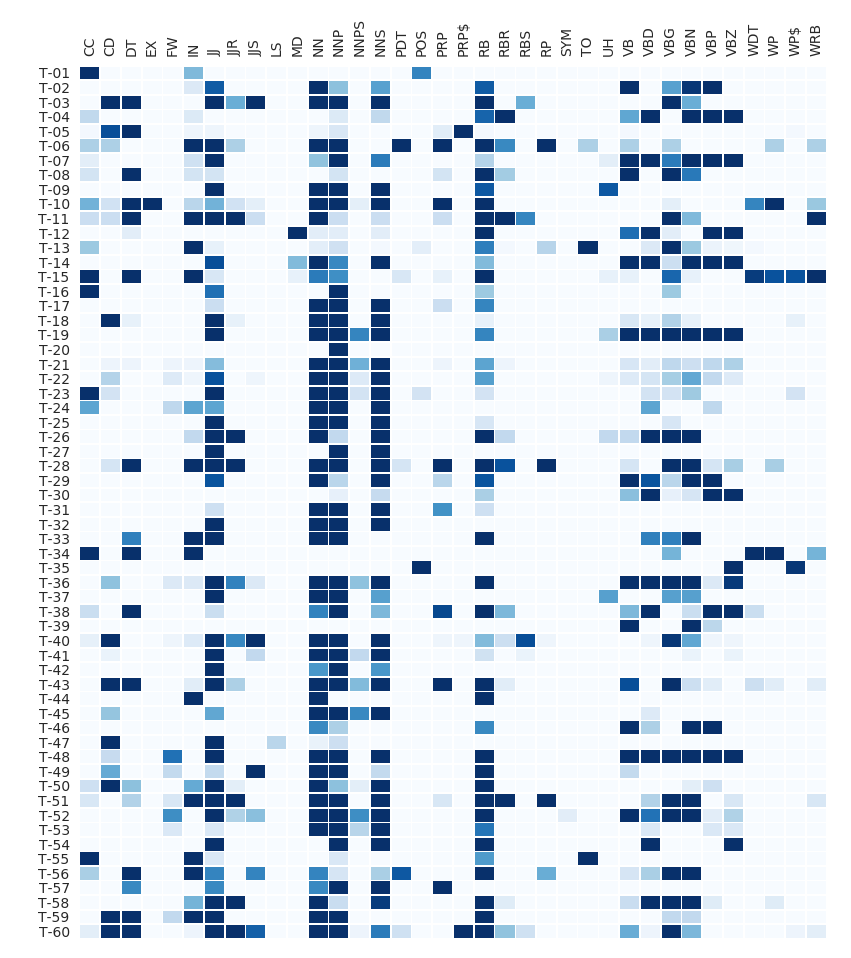}
    \includegraphics[scale=0.48]{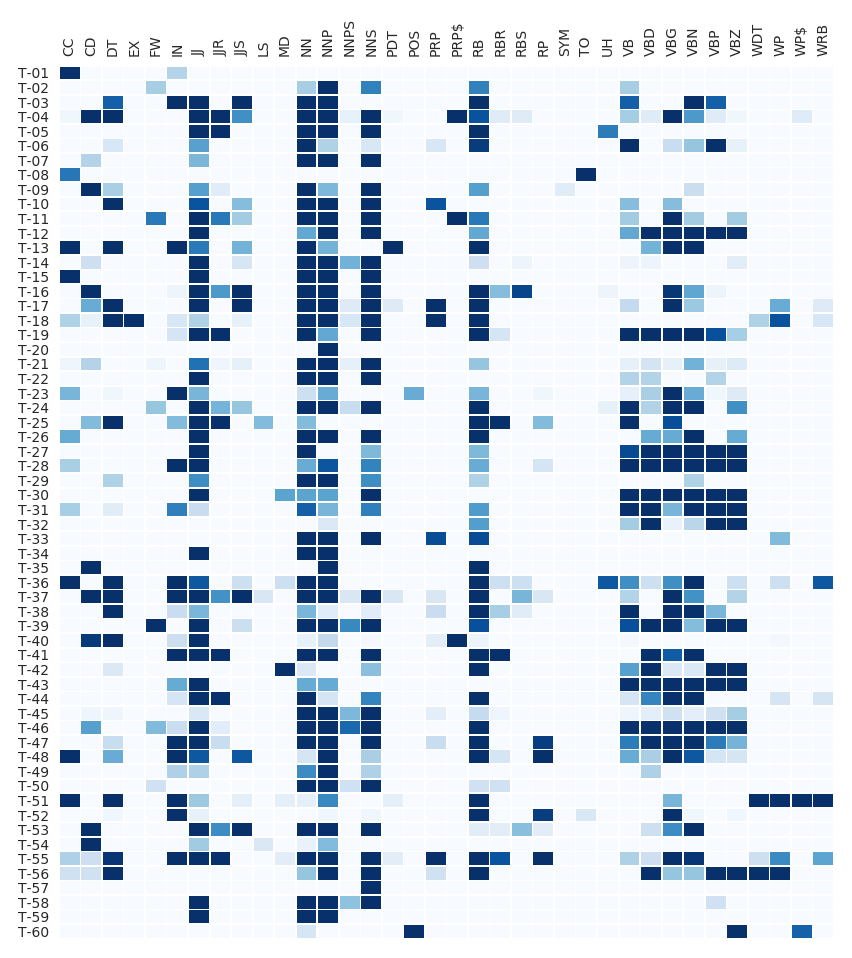}
    \caption{Preterminal alignment to part-of-speech tags for the compound PCFG (top) and the neural PCFG (bottom).}
    \label{fig:pos}
\end{figure*}
\subsection{Nonterminal/Preterminal Alignments}\label{lab:align}
Figure~\ref{fig:pos} shows the part-of-speech alignments and Table~\ref{tab:labels}  shows the nonterminal label alignments for the compound PCFG/neural PCFG.

\subsection{Subtree Analysis}\label{lab:subtrees}
Table~\ref{tab:pca3} lists more examples of constituents within each subtree as the top principical component is varied. Due to data sparsity, the subtree analysis is performed on the full dataset. See section~\ref{sec:analysis} for more details.

\begin{table*}[ht]
    \centering
    \tiny
    \begin{tabular}{l l}
    
    \toprule
    \multicolumn{2}{c}{(NT-13 (T-12 $w_1$) (NT-25 (T-39 $w_2$) (T-58 $w_3$)))} \vspace{1mm} \\ 
\textsf{ would be irresponsible} & \textsf{has been growing} \\
\textsf{ could be delayed} & \textsf{'ve been neglected} \\
\textsf{ can be held} & \textsf{had been made} \\
\textsf{ can be proven} & \textsf{had been canceled} \\
\textsf{ could be used} & \textsf{have been wary} \\
\midrule
\multicolumn{2}{c}{(NT-04 (T-13 $w_1$) (NT-12 (T-60 $w_2$) (NT-18 (T-60 $w_3$) (T-21 $w_4$))))} \vspace{1mm} \\ 
\textsf{ of federally subsidized loans} \hspace{55mm} &   \textsf{in fairly thin trading} \\
\textsf{ of criminal racketeering charges} & \textsf{in quiet expiration trading} \\
\textsf{ for individual retirement accounts} & \textsf{in big technology stocks} \\
\textsf{ without prior congressional approval} & \textsf{from small price discrepancies} \\
\textsf{ between the two concerns} & \textsf{by futures-related program buying} \\
\midrule

\multicolumn{2}{c}{(NT-04 (T-13 $w_1$) (NT-12 (T-05 $w_2$) (NT-01 (T-18 $w_3$) (T-25 $w_4$))))} \vspace{1mm} \\ 
\textsf{ by the supreme court} & \textsf{in a stock-index arbitrage} \\
\textsf{ of the bankruptcy code} & \textsf{as a hedging tool} \\
\textsf{ to the bankruptcy court} & \textsf{of the bond market} \\
\textsf{ in a foreign court} & \textsf{leaving the stock market} \\
\textsf{ for the supreme court} & \textsf{after the new york} \\
\midrule
\multicolumn{2}{c}{(NT-12 (NT-20 (NT-20 (T-05 $w_1$) (T-40 $w_2$)) (T-40 $w_3$)) (T-22 $w_4$))} \vspace{1mm} \\ 
\textsf{ a syrian troop pullout} & \textsf{the frankfurt stock exchange} \\
\textsf{ a conventional soviet attack} & \textsf{the late sell programs} \\
\textsf{ the house-passed capital-gains provision} & \textsf{a great buying opportunity} \\
\textsf{ the official creditors committee} & \textsf{the most active stocks} \\
\textsf{ a syrian troop withdrawal} & \textsf{a major brokerage firm} \\
\midrule
\multicolumn{2}{c}{(NT-21 (NT-22 (NT-20 (T-05 $w_1$) (T-40 $w_2$)) (T-22 $w_3$)) (NT-13 (T-30 $w_4$) (T-58 $w_5$)))} \vspace{1mm} \\ 
\textsf{ the frankfurt market was mixed} & \textsf{the gramm-rudman targets are met} \\
\textsf{ the u.s. unit edged lower} & \textsf{a private meeting is scheduled} \\
\textsf{ a news release was prepared} & \textsf{the key assumption is valid} \\
\textsf{ the stock market closed wednesday} & \textsf{the budget scorekeeping is completed} \\
\textsf{ the stock market remains fragile} & \textsf{the tax bill is enacted} \\
\midrule
\multicolumn{2}{c}{(NT-03 (T-07 $w_1$) (NT-19 (NT-20 (NT-20 (T-05 $w_2$) (T-40 $w_3$)) (T-40 $w_4$)) (T-22 $w_5$)))} \vspace{1mm} \\ 
\textsf{ have a high default risk} & \textsf{rejected a reagan administration plan} \\
\textsf{ have a lower default risk} & \textsf{approved a short-term spending bill} \\
\textsf{ has a strong practical aspect} & \textsf{has an emergency relief program} \\
\textsf{ have a good strong credit} & \textsf{writes the hud spending bill} \\
\textsf{ have one big marketing edge} & \textsf{adopted the underlying transportation measure} \\
\midrule
\multicolumn{2}{c}{(NT-13 (T-12 $w_1$) (NT-25 (T-39 $w_2$) (NT-23 (T-58 $w_3$) (NT-04 (T-13 $w_4$) (T-43 $w_5$)))))} \vspace{1mm} \\ 
\textsf{ has been operating in paris} & \textsf{will be used for expansion} \\
\textsf{ has been taken in colombia} & \textsf{might be room for flexibility} \\
\textsf{ has been vacant since july} & \textsf{may be built in britain} \\
\textsf{ have been dismal for years} & \textsf{will be supported by advertising} \\
\textsf{ has been improving since then} & \textsf{could be used as weapons} \\
\midrule
\multicolumn{2}{c}{(NT-04 (T-13 $w_1$) (NT-12 (NT-06 (NT-20 (T-05 $w_2$) (T-40 $w_3$)) (T-22 $w_4$)) (NT-04 (T-13 $w_5$) (NT-12 (T-18 $w_6$) (T-53 $w_7$)))))} \vspace{1mm} \\ 
\textsf{ for a health center in south carolina } \hspace{55mm} & \textsf{with an opposite trade in stock-index futures} \\
\textsf{ by a federal jury in new york} & \textsf{from the recent volatility in financial markets} \\
\textsf{ of the appeals court in new york} & \textsf{of another steep plunge in stock prices} \\
\textsf{ of the further thaw in u.s.-soviet relations} & \textsf{over the past decade as pension funds} \\
\textsf{ of the service corps of retired executives} & \textsf{by a modest recovery in share prices} \\
\midrule
\multicolumn{2}{c}{(NT-10 (T-55 $w_1$) (NT-05 (T-02 $w_2$) (NT-19 (NT-06 (T-05 $w_3$) (T-41 $w_4$)) (NT-04 (T-13 $w_5$) (NT-12 (T-60 $w_6$) (T-21 $w_7$))))))} \vspace{1mm} \\ 
\textsf{ to integrate the products into their operations} & \textsf{to defend the company in such proceedings} \\
\textsf{ to offset the problems at radio shack} & \textsf{to dismiss an indictment against her claiming} \\
\textsf{ to purchase one share of common stock} & \textsf{to death some N of his troops} \\
\textsf{ to tighten their hold on their business} & \textsf{to drop their inquiry into his activities} \\
\textsf{ to use the microprocessor in future products} & \textsf{to block the maneuver on procedural grounds} \\
\midrule
\multicolumn{2}{c}{(NT-13 (T-12 $w_1$) (NT-25 (T-39 $w_2$) (NT-23 (T-58 $w_3$) (NT-04 (T-13 $w_4$) (NT-12 (NT-20 (T-05 $w_5$) (T-40 $w_6$)) (T-22 $w_7$))))))} \vspace{1mm} \\ 
\textsf{ has been mentioned as a takeover candidate} & \textsf{would be run by the joint chiefs} \\
\textsf{ has been stuck in a trading range} & \textsf{would be made into a separate bill} \\
\textsf{ had left announced to the trading mob} & \textsf{would be included in the final bill} \\
\textsf{ only become active during the closing minutes} & \textsf{would be costly given the financial arrangement} \\
\textsf{ will get settled in the short term} & \textsf{would be restricted by a new bill} \\
\midrule
\multicolumn{2}{c}{(NT-10 (T-55 $w$) (NT-05 (T-02 $w_1$) (NT-19 (NT-06 (T-05 $w_2$) (T-41 $w_3$)) (NT-04 (T-13 $w_4$) (NT-12 (T-60 $w_5$) (NT-18 (T-18 $w_6$) (T-53 $w_7$)))))))} \vspace{1mm} \\ 
\textsf{ to supply that country with other defense systems} & \textsf{to enjoy a loyalty among junk bond investors} \\
\textsf{ to transfer its skill at designing military equipment} & \textsf{to transfer their business to other clearing firms} \\
\textsf{ to improve the availability of quality legal service} & \textsf{to soften the blow of declining stock prices} \\
\textsf{ to unveil a family of high-end personal computers} & \textsf{to keep a lid on short-term interest rates} \\
\textsf{ to arrange an acceleration of planned tariff cuts} & \textsf{to urge the fed toward lower interest rates} \\
\midrule
\multicolumn{2}{c}{(NT-21 (NT-22 (T-60 $w_1$) (NT-18 (T-60 $w_2$) (T-21 $w_3$))) (NT-13 (T-07 $w_4$) (NT-02 (NT-27 (T-47 $w_5$) (T-50 $w_6$)) (NT-10 (T-55 $w_7$) (NT-05 (T-47 $w_8$) (T-50 $w_9$))))))} \vspace{1mm} \\ 
\textsf{ unconsolidated pretax profit increased N \% to N billion} & \textsf{amex short interest climbed N \% to N shares} \\
\textsf{ its total revenue rose N \% to N billion} & \textsf{its pretax profit rose N \% to N million} \\
\textsf{ total operating revenue grew N \% to N billion} & \textsf{its pretax profit rose N \% to N billion} \\
\textsf{ its group sales rose N \% to N billion} & \textsf{fiscal first-half sales slipped N \% to N million} \\
\textsf{ total operating expenses increased N \% to N billion} & \textsf{total operating expenses increased N \% to N billion} \\
\bottomrule

    \end{tabular}
    \vspace{-3mm}
    \caption{For each subtree (shown at the top of each set of examples), we perform PCA on the variational posterior mean vectors that are associated with that particular subtree and take the top principal component. We then list the top 5 constituents that had the lowest (left) and highest (right) principal component values.}
    \label{tab:pca3}
    \vspace{-7mm}
\end{table*}

      \end{document}